%% file: main.tex
\documentclass[runningheads]{llncs}

\usepackage[mobile]{eccv}

\usepackage{eccvabbrv}

\usepackage{graphicx}
\usepackage{booktabs}

\usepackage[accsupp]{axessibility}  %

\usepackage{hyperref}

\usepackage{orcidlink}

\usepackage{algorithm}
\usepackage{algpseudocode}
\usepackage{adjustbox}

\AtBeginDocument{%
    \setcounter{secnumdepth}{3} %
    \crefname{subsubsection}{subsubsection}{subsubsections}
    \Crefname{subsubsection}{Subsubsection}{Subsubsections}
}

\newcommand{\equalcontrib}{\textsuperscript{\ensuremath{\star}}}
\newcommand{\equalcontribfootnote}{%
  \begingroup
  \renewcommand{\thefootnote}{\ensuremath{\star}}%
  \footnotetext[0]{Equal contribution.}%
  \endgroup
}

\usepackage{array}
\usepackage{multirow}

\newlength{\epecolw}
\newcolumntype{E}{>{\centering\arraybackslash}p{\epecolw}}

\newcommand{\epegroup}[1]{%
  \multicolumn{3}{c}{\makebox[\dimexpr3\epecolw+4\tabcolsep\relax][c]{#1}}}

\begin{document}

\title{InFlux++: Real and Synthetic Data for Estimating Dynamic Camera Intrinsics}

\titlerunning{InFlux++}

\author{Erich Liang\orcidlink{0000-0001-7414-0194} \and
Caleb Kha-Uong\equalcontrib\orcidlink{0009-0000-9093-3821} \and
Chinmaya Saran\equalcontrib\orcidlink{0009-0008-1302-5481} \and
Sreemanti Dey\equalcontrib\orcidlink{0000-0003-2428-3538} \and
David W. Liu\orcidlink{0000-0002-0605-3045} \and
Junhan Ouyang\orcidlink{0009-0000-5049-5679} \and
Benjamin Zhou\orcidlink{0009-0001-0376-1513} \and
Jia Deng\orcidlink{0000-0001-9594-4554}
}

\authorrunning{E.~Liang et al.}

\institute{Princeton University, Princeton NJ 08544, USA\\
\email{\{erliang,ck2867,cs4046,sd9968,dl3533,\\
harryouyang,bz2883,jiadeng\}@princeton.edu}
}

\maketitle
\equalcontribfootnote

\begin{abstract}
    Camera intrinsics play a vital role in recovering 3D structure from 2D video. However, most 3D algorithms assume that intrinsics remain fixed throughout input video, an assumption that frequently fails for real-world in-the-wild videos. Consequently, estimating per-frame intrinsics from RGB images is critical for enabling 3D methods to operate robustly on dynamic intrinsics videos. Previously, InFlux has contributed to this research direction by establishing the first real-world benchmark with per-frame ground truth intrinsics for dynamic intrinsics videos, enabling systematic evaluation of algorithms for this task. Nevertheless, existing methods remain inaccurate due to two key obstacles: (i) training data for this task is scarce and lacks diversity in camera intrinsics; and (ii) existing benchmarks, including InFlux, are limited in scene and camera motion diversity, making it difficult to properly evaluate method performance. To address both gaps, we present InFlux++, consisting of two components. InFlux++ Synth is a large-scale procedurally generated synthetic video dataset with 441K+ annotated frames from 1841 high-resolution videos, providing accurate per-frame ground truth intrinsics for training dynamic intrinsics prediction models; a subset additionally includes per-frame camera pose, depth, and surface normals. The videos feature rich intrinsics diversity through changes in camera zoom and focus over time, as well as dynamic objects and realistic rendering effects such as lens distortion and defocus blur. InFlux++ Real is a large-scale real-world benchmark that extends InFlux with 514K+ newly captured frames across 334 high-resolution videos, spanning a wider range of scenes and camera motions. Finetuning existing intrinsics prediction methods on InFlux++ Synth consistently improves focal length estimation across both InFlux++ Real and InFlux, suggesting that synthetic supervision is a promising direction for RGB-based intrinsics prediction. For the dataset, benchmark, code, videos, submission instructions, and live leaderboard, please visit \url{https://influx.cs.princeton.edu/}.

\end{abstract}

\section{Introduction}
\label{sec:intro}

Camera intrinsics are fundamental to many real-world 3D systems, as they define the geometric mapping between 3D coordinates and the 2D image. Example applications include robotic depth perception and digital overlays in AR/VR.

\begin{figure}[t]
  \centering
  \includegraphics[width=\linewidth]{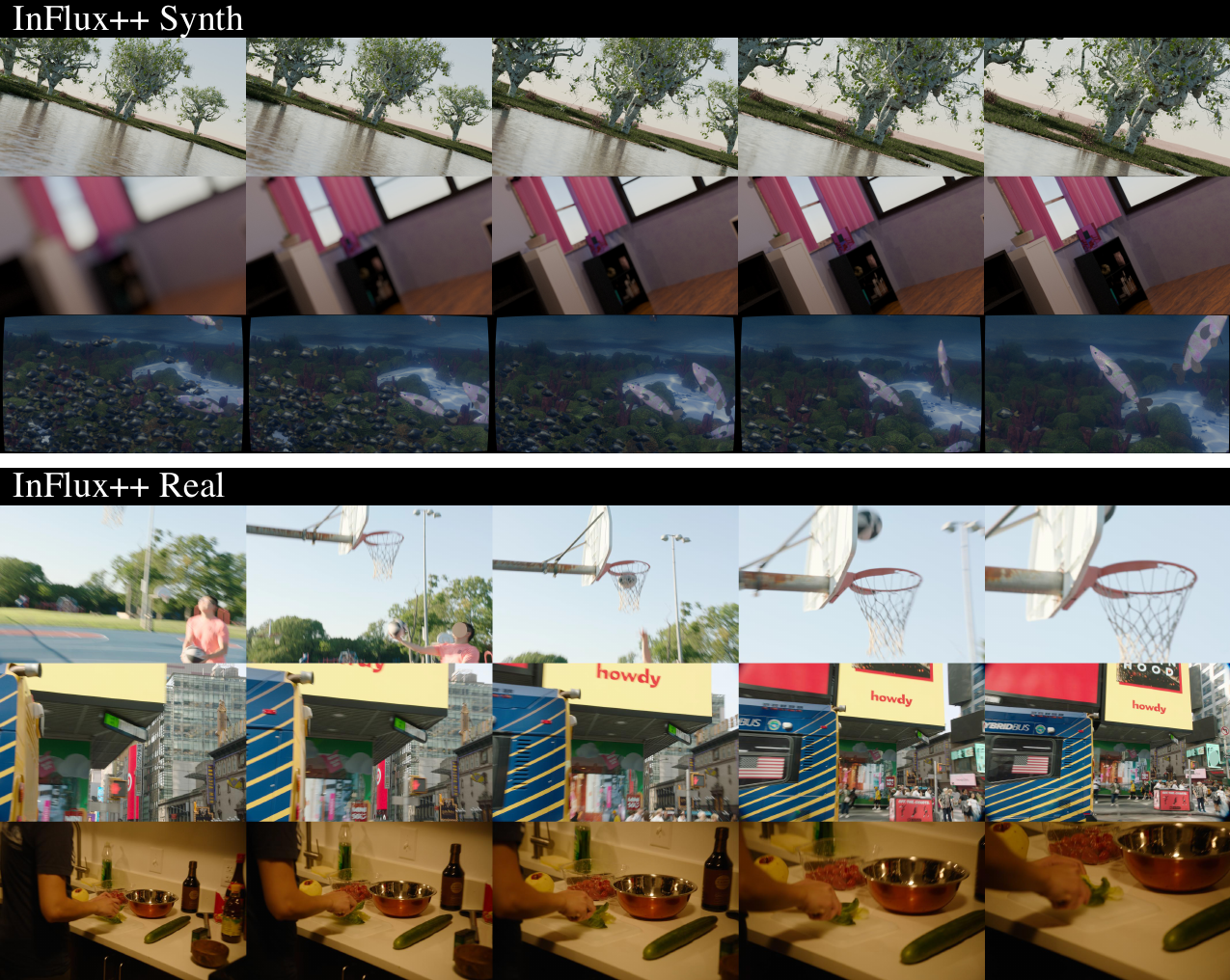}
  \caption{A gallery of InFlux++, our data suite for dynamic intrinsics prediction. The top three rows show videos from InFlux++ Synth, a synthetic dataset with 441K+ frames across 1841 videos, featuring diverse intrinsics from changing zoom and focus, varying lens distortion, dynamic objects, and realistic optical effects such as defocus blur. The bottom three rows show videos from InFlux++ Real, a real-world dynamic intrinsics benchmark with 514K+ frames across 334 high-resolution videos. Compared to InFlux~\cite{influx}, it captures a wider range of indoor and outdoor scenes, natural daily activities, and richer camera motions with more translational motion.}
  \label{fig:gallery}
\end{figure}

Many modern 3D algorithms~\cite{colmap, droid-slam, DPVO, orb-slam2, NeRF, GS} assume that camera intrinsics remain constant throughout video input, but real-world in-the-wild footage often violates this assumption. Intrinsics can vary during capture due to changes in zoom or focus distance, which alter the optical configuration of the lens.

As a result, accurately predicting per-frame camera intrinsics from RGB frames in dynamic intrinsics video is critical for extending standard 3D algorithms to real-world videos. Many 3D algorithms rely heavily on camera intrinsics for downstream geometric computations and assume that these parameters remain constant throughout the input video. When intrinsics change during capture, this constant-intrinsics assumption leads to incorrect calibration values being used throughout the pipeline, degrading downstream computations. However, if intrinsics can be estimated accurately for each frame, they could be directly supplied to existing 3D algorithms, enabling reliable operation on dynamic intrinsics video without requiring fundamental architectural changes.

Previously, InFlux~\cite{influx} has contributed to this direction of intrinsics prediction for dynamic intrinsics video by establishing the first real-world benchmark for this task, enabling systematic evaluation of prediction methods. However, state of the art (SOTA) methods such as~\cite{anycalib} still incur large intrinsics prediction errors. When measuring prediction error via endpoint error (EPE), which quantifies the 2D distance between projections of the same 3D points under ground truth and predicted intrinsics, only 34.1\% of projected points fall within 50 pixels of their ground truth locations. We identify two main obstacles in the field that limit intrinsics prediction performance.

\begin{figure}[t]
    \centering
    \includegraphics[width=\linewidth]{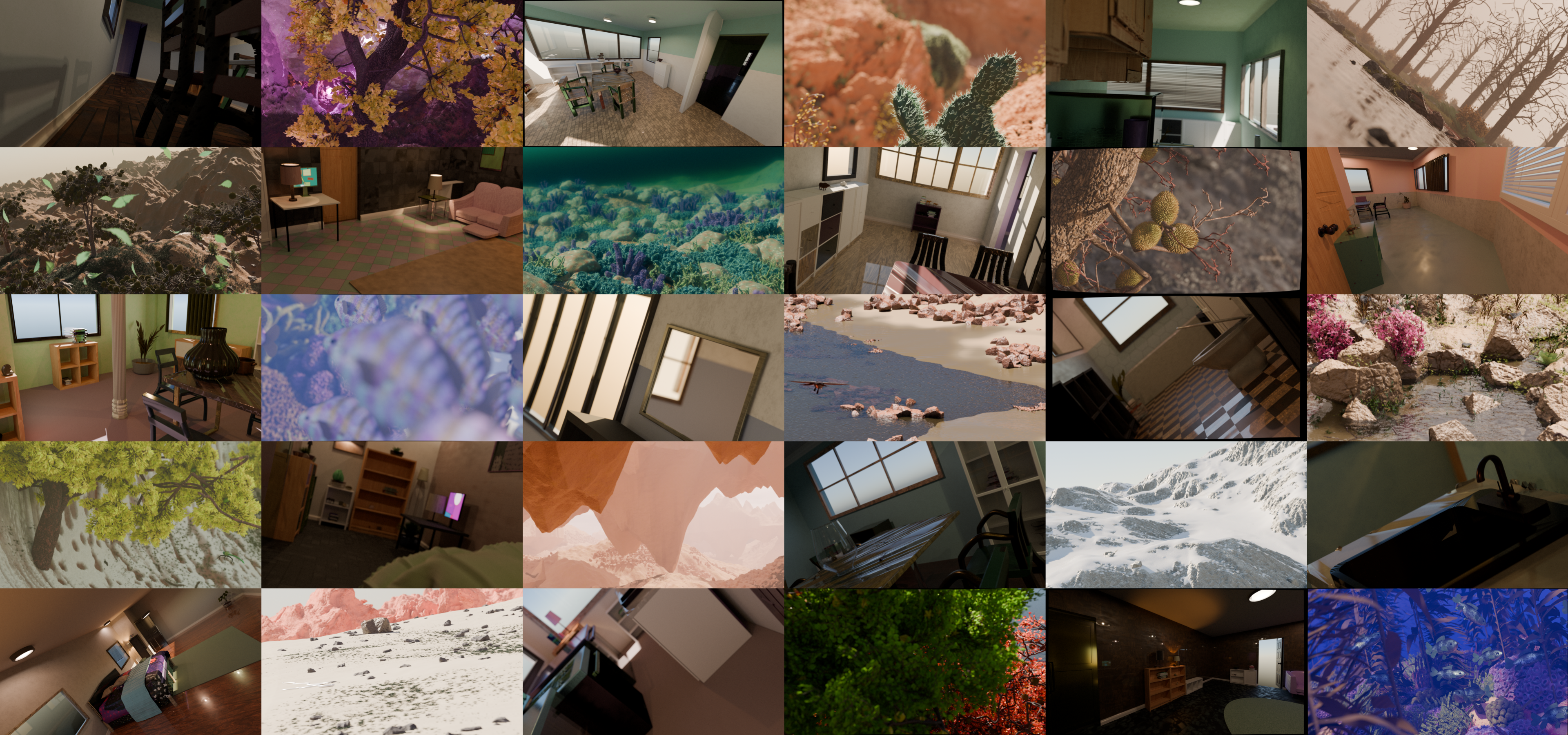}
    \caption{Gallery of InFlux++ Synth's scene diversity. The dataset includes a wide range of procedurally generated indoor and nature environments with diverse lighting, materials, and scene composition. Indoor scenes feature varied room structures and non-Lambertian surfaces, while nature scenes span settings like forests, deserts, mountains, and underwater environments with rich terrain, vegetation, water bodies, and dynamic elements such as animals, falling leaves, and water motion.}
    \label{fig:synth_diversity}
\end{figure}

The first obstacle to improving dynamic intrinsics prediction accuracy is the scarcity and limited diversity of training data for this task. Although many real-world datasets~\cite{KITTI, EuRoC, TUM-RGBD, ETH-3D} provide camera intrinsics, the intrinsics typically remain constant throughout each dataset. Panorama datasets~\cite{geocalib, anycalib} can be used to generate image crops with custom focal lengths, but they cannot capture the realistic temporal changes in intrinsics or camera motion found in video, and their accuracy can be affected by manufacturing imperfections in $360^\circ$ cameras or stitching artifacts introduced during panorama construction. Some synthetic datasets such as~\cite{TartanAir} provide videos with per-frame intrinsics, but most do not vary the intrinsics across frames or videos. Among the few that do,~\cite{bedlamv2} varies focal length but does not model lens distortion or defocus blur, limiting its usefulness for training models for dynamic intrinsics video prediction.

The second obstacle to improving dynamic intrinsics prediction accuracy is the limited scene and camera motion diversity in existing benchmarks for this task. In terms of scene diversity, InFlux~\cite{influx} outdoor coverage consists mostly of shots of outdoor university campus areas or natural landmarks, with relatively few shots in busy urban environments. Indoor coverage is limited to office or workspace settings, or empty laboratory spaces featuring staged skits, which do not capture the range of natural daily activities found in real-world footage. Regarding camera motion, InFlux predominantly features rotational motion, with little camera translation or parallax. This lack of varied camera motion may not adequately reflect the camera movement found in real-world videos.

\begin{figure}[t]
  \centering
  \includegraphics[width=\linewidth]{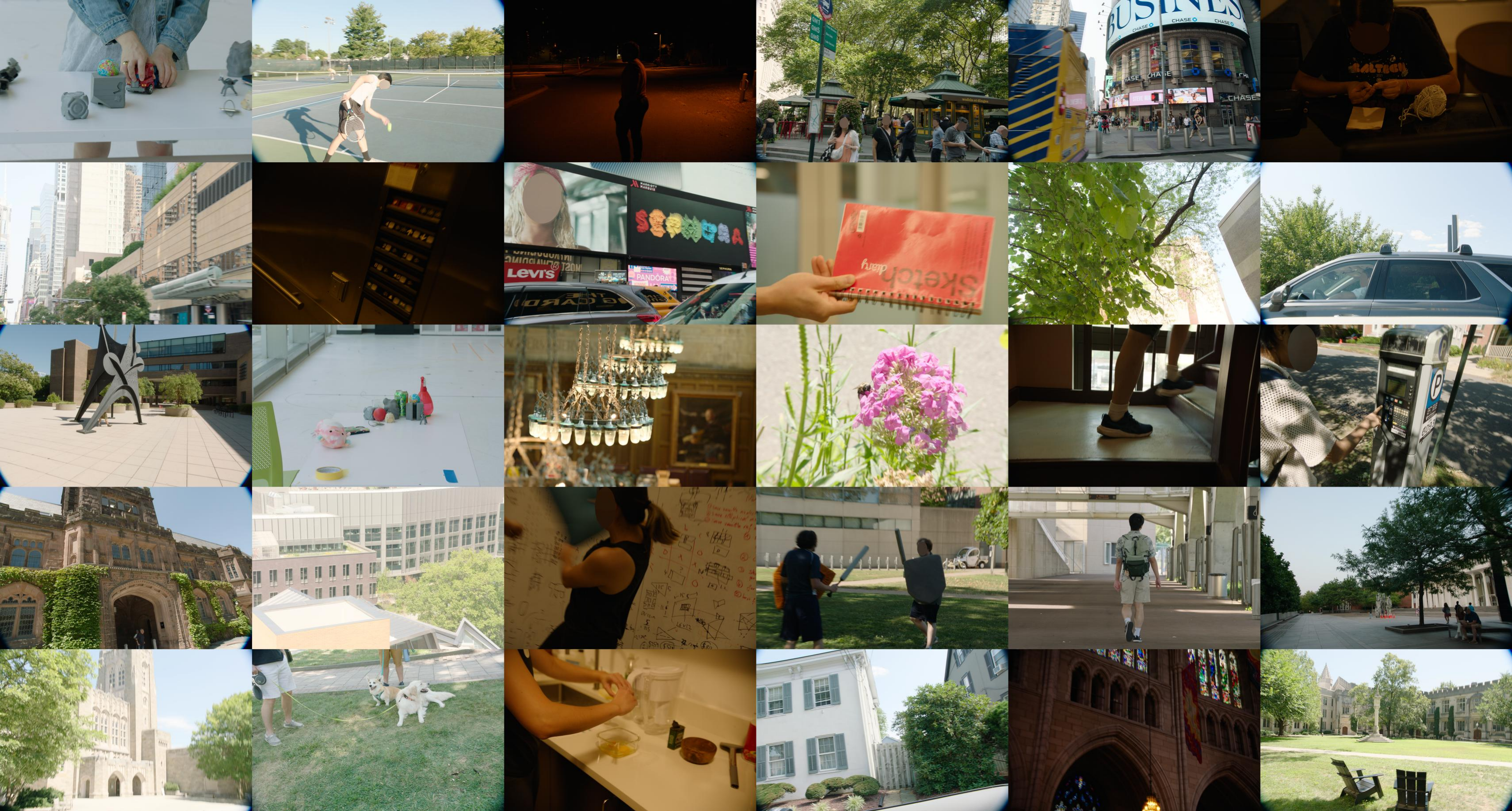}
  \caption{Gallery of InFlux++ Real's scene diversity. Our benchmark contains a wide range of environments, activities, and viewpoints.}
  \label{fig:real_diversity}
\end{figure}

To address these limitations, we introduce InFlux++, a comprehensive data suite for dynamic intrinsics prediction, comprising a large-scale synthetic training set and an extended real-world benchmark. InFlux++ Synth is a procedurally generated synthetic video dataset with 441K+ annotated frames from 1841 high-resolution videos, designed for supervised training of dynamic intrinsics prediction models. Building on Infinigen~\cite{infinigen, infinigen2024indoors}, we extend the renderer to generate accurate per-frame ground truth intrinsics with high diversity in camera parameters, achieved by varying both zoom and focus throughout each video. A subset of the dataset also provides camera pose, depth, and surface normals. The videos include dynamic objects, realistic optical effects such as lens distortion and defocus blur, and capture dynamic video sequences rather than isolated images, addressing the limited intrinsics diversity and lack of frame-to-frame variation in prior datasets. InFlux++ Real extends the original InFlux benchmark with 514K+ newly captured frames across 334 high-resolution videos, directly addressing prior limitations in scene and motion diversity. Compared to InFlux, it includes substantially more outdoor urban environments, a broader range of indoor settings beyond offices and laboratory scenes, and a greater emphasis on naturally occurring real-world activities. It also features more camera translation than InFlux, producing greater parallax and better reflecting real-world camera motion. Together, these resources provide both the training diversity and benchmark coverage needed to systematically develop and evaluate robust dynamic intrinsics prediction methods. A gallery of InFlux++ can be found in \cref{fig:gallery}.

Using InFlux++ Real and InFlux as evaluation data, we assess the performance of existing intrinsics prediction baselines~\cite{anycalib, colmap, droidcalib, geocalib, perspectivefields, unidepthv2, wildcamera}. We further finetune AnyCalib~\cite{anycalib}, the current SOTA method for dynamic intrinsics prediction, on InFlux++ Synth. Evaluation results show that while dynamic intrinsics prediction remains challenging, finetuning on InFlux++ Synth yields improved focal length estimation across both InFlux++ Real and InFlux. These results suggest that synthetic supervision is a promising direction for RGB-based dynamic intrinsics prediction. In summary, our contributions are:

\begin{itemize}
    \item We introduce InFlux++ Synth, a procedurally generated synthetic video dataset with accurate and dynamic intrinsics spanning 441K+ frames across 1841 videos. A subset also includes pose, depth, and normals annotations. The videos feature diverse intrinsics from changes in zoom and focus, varying lens distortion, dynamic objects, and realistic optical effects like defocus blur.

    \item We introduce InFlux++ Real, a real-world dynamic intrinsics benchmark extending InFlux with 514K+ frames across 334 high-resolution videos. Compared to InFlux, it includes substantially more outdoor urban environments, a broader range of indoor settings, more naturally occurring real-world activities, and richer camera trajectories featuring more translational motion.
    
    \item We demonstrate that finetuning the SOTA method AnyCalib~\cite{anycalib} on InFlux++ Synth leads to improved focal length estimation on InFlux++ Real and InFlux, suggesting that synthetic supervision is a promising direction for RGB-based dynamic intrinsics prediction.
\end{itemize}

\section{Preliminaries}
In this paper, we adopt similar camera and lens terminology established in~\cite{influx}:
\\
\\
\textbf{Camera Focal Length (CFL)} is the distance from a camera's optical center to its imaging plane along the optical axis. CFL corresponds to the $f_x$ and $f_y$ entries of the camera intrinsics matrix when measured in pixels.
\\
\\
\textbf{Lens Focal Length (LFL)} is the CFL of a camera when it is focused at infinity~\cite{lensdef}. Effective CFL often differs from the LFL for real-world images.
\\
\\
\textbf{Lens to Object Distance (LTO)} is the distance from the camera's optical center to the object in focus, measured along the optical axis.
\\
\\
\textbf{Focus Distance (FD)} is the distance along the optical axis from the camera's sensor plane to the object in focus. It is the sum of CFL and LTO.
\\
\\
\textbf{Zoom Lenses} can produce dynamic camera intrinsics during video filming via mechanical shifts in the zoom ring, which controls LFL, and the focus ring, which controls FD. Some specialized zoom lenses report per-frame LFL and FD as lens metadata, which together fully characterize the optical state of the lens~\cite{practical_zoom}.

\section{Related Work}

\subsection{Training Data for Dynamic Intrinsics Prediction}

\textbf{Real-World Training Datasets.} Popular datasets like~\cite{TUM-RGBD, EuRoC, KITTI} provide camera intrinsics, but are captured with pre-calibrated camera rigs with fixed intrinsics, making them unsuitable for supervising dynamic intrinsics prediction. \cite{ETH-3D} records videos using a small set of different lenses. Although this introduces variation in camera intrinsics, the variation occurs only across different videos---intrinsics still remain fixed within each individual video sequence, making the dataset unsuitable for supervising dynamic per-frame intrinsics prediction. \cite{megadepth} provides COLMAP-derived intrinsics for Internet photo collections~\cite{colmap}, but lacks temporal video structure and relies on reconstruction-derived pseudo ground truth. \cite{lamar, lamaria} include time-varying intrinsics for AR device images, but the changes are incidental and small, arising from effects like autofocus or changes in device temperature. Panorama-based datasets such as~\cite{geocalib, anycalib} construct images with varying intrinsics by cropping $360^\circ$ panoramas. While this introduces large changes in intrinsics, the images lack temporal correspondence and rely on ideal panorama geometry, making them susceptible to stitching artifacts and lens imperfections. In contrast, InFlux++ Synth provides high-resolution synthetic video sequences with changing intrinsics and perfectly accurate per-frame annotations, enabling effective supervision for dynamic intrinsics prediction.
\\
\\
\textbf{Synthetic Training Datasets.} Synthetic datasets inherently provide perfectly accurate intrinsics, thanks to privileged knowledge of the rendering camera, but few are suitable for supervising dynamic intrinsics prediction. \cite{TartanAir, scenenet, interiornet} use a fixed focal length for all renders, so they cannot be effectively used for training dynamic intrinsics prediction models. \cite{bedlamv2} provides videos with varying focal length, but it only uses the simple pinhole camera model for rendering. As a result, its videos lack defocus blur, which occurs in real camera systems and could serve as a useful training signal for intrinsics prediction. Its videos also lack lens distortion, which reduces the diversity of intrinsics. In contrast, InFlux++ Synth uses the thin lens model for video rendering, enabling realistic rendering effects like defocus blur. It also includes a data loader that applies customizable Brown-Conrady~\cite{brown} image distortion, thus providing diverse yet accurate intrinsics for realistic videos to support training of dynamic intrinsics prediction models.

\subsection{Benchmarks for Dynamic Intrinsics Prediction}

InFlux~\cite{influx} is the closest existing benchmark, offering per-frame ground truth intrinsics across varied cameras and scenes. However, its scene coverage is narrow, skewing toward university campuses and natural landmarks for outdoor scenes, and office or laboratory spaces with staged subjects for indoor scenes. InFlux's camera motion is also predominantly rotational, with little translation or parallax. In contrast, our benchmark InFlux++ Real covers a broader range of environments, including busy urban settings, everyday indoor activities like cooking and laundry, and diverse outdoor scenes such as sports and recreation. InFlux++ Real features longer videos with subjects performing natural everyday activities such as household chores or playing basketball. Camera motion includes more translational movement---including vehicle-mounted capture---resulting in richer parallax and a more faithful reflection of the diversity of real-world videos.

\section{Methodology}

In this section, we describe the construction of InFlux++, which consists of two parts: InFlux++ Synth, a large-scale synthetic video dataset with accurate per-frame intrinsics, realistic optical effects, dynamic objects, and pose, depth, and normals for a subset of frames; and InFlux++ Real, a real-world benchmark that extends InFlux~\cite{influx} with diverse scenes, activities, and camera motions.

InFlux++ Synth (\cref{subsec:influxpp_synth}) builds on Infinigen~\cite{infinigen, infinigen2024indoors} (\cref{subsubsec:synth_rendering}) to generate diverse indoor and nature scenes with accurate per-frame annotations, adding camera motion (\cref{subsubsec:synth_motion}), dynamic intrinsics (\cref{subsubsec:synth_computation}, \cref{subsubsec:synth_walk}), and lens distortion with other augmentations (\cref{subsubsec:synth_distort}) to produce a diverse (\cref{subsubsec:synth_diversity}) dataset suitable for training dynamic intrinsics prediction models.

InFlux++ Real (\cref{subsec:influxpp_real}) extends InFlux with 514K+ frames across 334 high-resolution videos. Ground truth intrinsics are obtained via InFlux's lens calibration methodology applied to our lenses, with modifications to the procedure for large field of view (FOV) spatial footprint calibration experiments (\cref{subsubsec:real_calib}). After lens calibration, we collect videos and blur frames containing faces or license plates (\cref{subsubsec:real_blur}) for privacy. Our benchmark videos are diverse (\cref{subsubsec:real_diversity}), covering a wider range of scenes and camera motions than InFlux.

\subsection{InFlux++ Synth}
\label{subsec:influxpp_synth}

\subsubsection{Procedural Scene Generation and Renderer.}
\label{subsubsec:synth_rendering}

We build InFlux++ Synth on top of Infinigen~\cite{infinigen, infinigen2024indoors}, a procedural scene generator and renderer, for two main reasons. First, Infinigen allows the creation of a virtually limitless variety of indoor and nature scenes, with diverse lighting, materials, and scene composition. Its layout optimizer and tuned parameter distributions ensure scene realism without any manual design, thus providing scene diversity critical for constructing a training dataset that generalizes well across environments.

Second, because Infinigen renders synthetic videos via Blender, we have full access to underlying camera and scene information, including the rendering camera's intrinsics. This allows us to generate diverse video sequences with perfect per-frame ground truth intrinsics even in the presence of dynamic objects, providing a level of accurate supervision that would be infeasible with real-world captures. We also provide per-frame camera pose, depth, and surface normals for a subset of the dataset; see our supplement for additional details. These capabilities make InFlux++ Synth well-suited for training models to predict dynamic intrinsics in varied and realistic scenarios.

\subsubsection{Camera Motion.}
\label{subsubsec:synth_motion}

To obtain varied camera motion, we build on the Rapidly-Exploring Random Trees (RRT) camera trajectory generator provided by Infinigen~\cite{infinigen}. After coarse scene geometry is generated, the planner samples collision-free positional waypoints, assigns camera orientations, and connects the resulting pose keyframes using Bézier interpolation. This results in natural camera motion with smooth translational and rotational motion.

To favor visually informative views with richer geometric structure, we add a surface normal and depth-based criterion that rejects camera orientations that produce viewpoints dominated by a single coherent surface, such as a blank wall or featureless terrain. We also strengthen orientation resampling and introduce local segment replanning, allowing RRT to recover from locally invalid viewpoints without discarding all previous motion. See our supplement for details.

\subsubsection{Per-Frame Intrinsics Parameterization.}
\label{subsubsec:synth_computation}

To produce realistic variation of camera intrinsics in our synthetic videos, we parameterize the CFL of the rendering camera using LFL and LTO, rather than directly setting CFL as Blender allows. This approach is inspired by real-world zoom lenses, where the camera intrinsics are fully parameterized by LFL and FD, which are physically adjusted via the zoom and focus rings, respectively.

To render a video with varying intrinsics, we vary the LFL and LTO values of each frame. For now, we focus on computing the CFL for a single frame. Given the LFL and LTO for a frame, we use the thin lens model to compute CFL as a function of these two parameters. This approach is consistent with Blender's Cycles rendering engine, which performs ray tracing based on the thin lens equation. Specifically, we have that:
\begin{align*}
    \frac{1}{\text{LFL}} &= \frac{1}{\text{CFL}} + \frac{1}{\text{LTO}}\\
    \text{CFL} &= \frac{\text{LFL} \cdot \text{LTO}}{\text{LTO} - \text{LFL}}
\end{align*}
By parameterizing CFL in this manner, each rendered frame's effective focal length changes in a physically meaningful manner. In addition, this formulation naturally produces lens breathing, a visual phenomenon observed in real-world cameras where the FOV can shift slightly when the LTO changes, even with a fixed LFL. This realistic visual effect does not occur by default in Blender, where CFL and LTO are independently set. By adding lens breathing to our video renders, we provide a potentially valuable supervisory signal for training models to predict dynamic intrinsics. See our supplement for details.

\subsubsection{Temporal Variation of LFL and LTO.}
\label{subsubsec:synth_walk}

Given our parameterization of CFL as a function of LFL and LTO, we now describe how we vary LFL and LTO over time to produce realistic changing intrinsics for our synthetic videos.

To vary LFL, we use a bounded random walk. Suppose the initial LFL value is $x_0$ at frame $t = 0$. We generate a target LFL for a future frame $t' > 0$ by adding a randomly sampled delta to $x_0$, ensuring the result stays within the allowed LFL bounds. The frame $t'$ itself is also chosen randomly within a specified range of steps ahead. This process is repeated sequentially across the video, producing a series of target LFL values at selected frames. We then use Bézier interpolation to smoothly fill in the LFL values for all intermediate frames between these targets, producing realistic smooth changes in LFL over time. For our renders, we bound LFL between $8~\text{mm}$ and $400~\text{mm}$. See \cref{alg:lfl_random_walk} for a summary of the bounded random walk method.

\begin{algorithm}[t]
\caption{Bounded Random Walk for LFL over Video Frames}
\label{alg:lfl_random_walk}
\begin{algorithmic}[1]
\State \textbf{Input:} 
\Statex \hspace{\algorithmicindent} $x_{\min}, x_{\max}$: LFL bounds
\Statex \hspace{\algorithmicindent} $\delta_{\max}$: maximum magnitude of LFL change between keyframes
\Statex \hspace{\algorithmicindent} $s_{\min}, s_{\max}$: min/max frames between keyframes
\Statex \hspace{\algorithmicindent} $T$: total number of video frames
\State Sample initial LFL $x_0 \sim \text{Uniform}(x_{\min}, x_{\max})$
\State Set $t \gets 0$, $x \gets x_0$ and record first keyframe $(t, x)$
\While{$t < T$}
    \Repeat
        \State Sample $\Delta x \sim \text{Uniform}(-\delta_{\max}, \delta_{\max})$
        \State $x_\text{new} \gets x + \Delta x$
    \Until{$x_\text{new} \in [x_{\min}, x_{\max}]$}
    \State Sample $\Delta t \sim \text{Uniform}(s_{\min}, s_{\max})$
    \State $t_\text{new} \gets t + \Delta t$
    \If{$t_\text{new} \geq T$}
        \State $t_\text{new} \gets T$ \Comment{ensure last frame is a keyframe}
    \EndIf
    \State Record keyframe: $(t_\text{new}, x_\text{new})$
    \State $t \gets t_\text{new}$, $x \gets x_\text{new}$
\EndWhile
\State Apply Bézier interpolation between keyframes to compute LFL for all other frames
\end{algorithmic}
\end{algorithm}

We follow a similar strategy to vary LTO. However, the visible scene content changes as the camera moves, so directly performing a bounded random walk with fixed limits on the raw LTO could produce low-quality frames in which the entire scene is out of focus. To address this, we estimate the minimum and maximum visible scene depths at each keyframe via ray tracing, denoted as $d_{\text{near}}$ and $d_{\text{far}}$, and parameterize the LTO as a fractional position $\alpha$ within this interval. The resulting LTO is computed as $d_{\text{focus}} = d_{\text{near}} + (d_{\text{far}} - d_{\text{near}})\alpha$, where $\alpha$ is constrained between 0 and 1 and changes via a bounded random walk. This formulation produces smooth temporal variation in LTO while ensuring that some portion of the scene remains in focus, even as camera motion alters the visible depth distribution. See our supplement for additional details.

\subsubsection{Lens Distortion and Augmentations.}
\label{subsubsec:synth_distort}

The videos produced by Infinigen are by default undistorted. To introduce lens distortion and other data augmentations useful for training dynamic intrinsics prediction models, we provide a data loader based on~\cite{geocalib, anycalib} that applies these effects on-the-fly.

For lens distortion, we adopt the Brown–Conrady~\cite{brown} model. To ensure realistic distortion, we first sample reasonable magnitudes of radial distortion at two control points on the edge of the image and then solve for $k_1, k_2$ values that produce these effects. We optionally sample small tangential $p_1, p_2$ values.  Our data loader can be further customized; see our supplement for details.

In addition to distortion, we also apply color transforms, resizing transforms, and noising operations following~\cite{anycalib}. However, we disable blurring, sharpening, and resolution changes. See our supplement for details.

\subsubsection{Data Diversity.}
\label{subsubsec:synth_diversity}

\begin{figure}[t]
    \centering
    \includegraphics[width=0.48\linewidth]{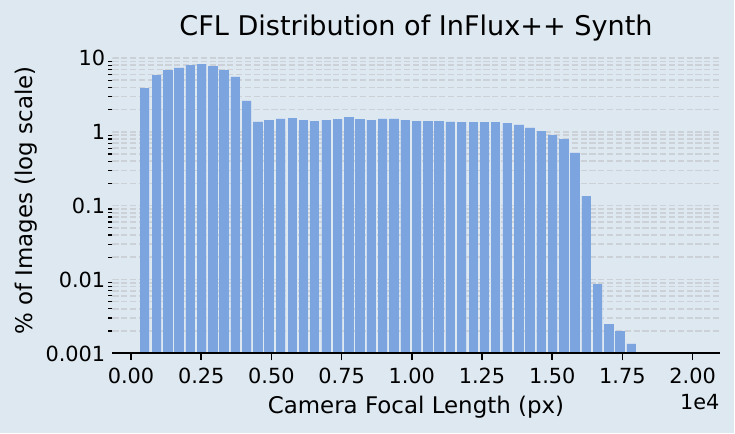}
    \includegraphics[width=0.48\linewidth]{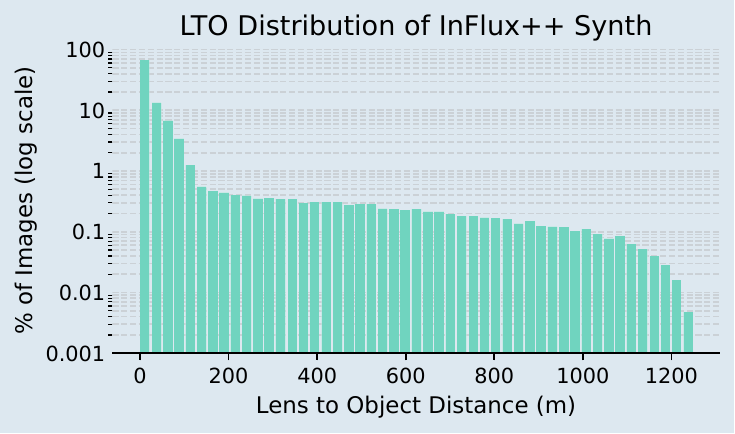}
    \includegraphics[width=0.48\linewidth]{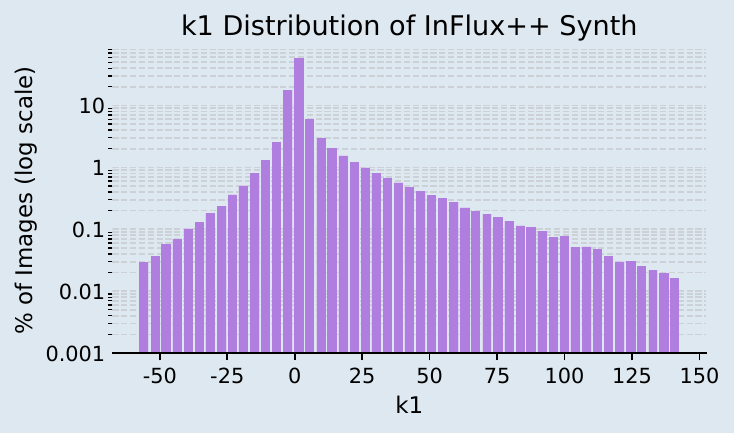}
    \includegraphics[width=0.48\linewidth]{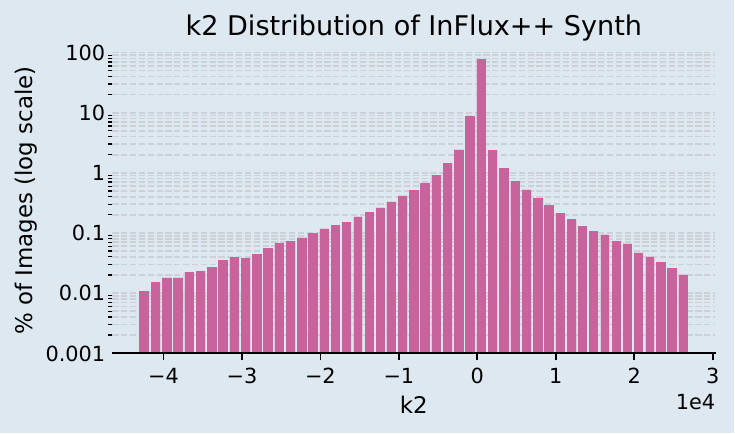}
    \caption{Histograms of per-frame intrinsics in InFlux++ Synth. Values of $k_1$ and $k_2$ are clipped at the 0.1\% and 99.9\% percentiles for visualization. Both are heavily centered around 0. LTO ranges roughly from 0.04~m to 12~m for indoor scenes and from 0.7~m upward for nature scenes. These value ranges arise naturally from scene geometry.}
    \label{fig:synth_hist_plots}
\end{figure}

We generate 441K+ images across 1841 videos, which are rendered at a resolution of $1280 \times 720$ and span 10 seconds at 24 FPS. Of these videos, 1040 are indoor scenes, and 801 are nature scenes.
\\
\\
\textbf{Camera Intrinsics Diversity.}
The bounded random walks used to vary LFL and LTO produce smooth, continuous variation in camera intrinsics throughout each sequence. 
As a result, downstream intrinsics parameterized by LFL and LTO, such as CFL, $k_1$, and $k_2$, exhibit broad coverage across their feasible ranges rather than clustering around a small set of discrete values. See \cref{fig:synth_hist_plots} for the empirical distributions of these parameters.
\\
\\
\textbf{Scene Diversity.} Our dataset includes a wide range of indoor environments, including living rooms, bedrooms, bathrooms, kitchens, office spaces, and more. These scenes exhibit realistic variation in layout, geometry, lighting conditions, and material properties. In particular, we include glossy and reflective surfaces as well as mirrors to capture challenging real-world optical effects.

Our nature scenes span diverse environments such as forests, deserts, caves, mountains, canyons, coasts, snowy arctic landscapes, coral reefs, and more. These environments are varied in scene layout, scale, geometry, texture, and visibility. Some scenes include dynamic objects, with animals such as birds, dragonflies, insects, snakes, and other terrestrial animals following collision-aware trajectories. Other scenes exhibit natural motion from water waves, ripples, and falling leaves. A qualitative gallery illustrating scene diversity is provided in \cref{fig:synth_diversity}.

\subsection{InFlux++ Real}
\label{subsec:influxpp_real}

To construct InFlux++ Real, we collect real-world videos with dynamic intrinsics and per-frame ground truth intrinsics annotations largely following the data collection procedure introduced in InFlux~\cite{influx}. Like InFlux, we utilize specialized zoom lenses that record lens metadata, including per-frame LFL and FD. Since these values uniquely determine the optical state of the lens, a one-time calibration process can be used to construct a lens-specific lookup table (LUT) mapping $(\text{LFL}, \text{FD})$ pairs to camera intrinsics. This allows per-frame intrinsics to be recovered during video capture without interrupting natural camera motion.

To construct a LUT for each lens, we perform camera calibration experiments over a wide grid of LFL and FD values. InFlux divides these experiments into two categories based on the FOV spatial footprint (FSF) of the camera, defined as the 3D region within the camera's FOV that is in focus. For small to medium FSF cases, we follow InFlux and perform calibration by moving boards of varying sizes throughout the camera's FSF. For large FSF cases, InFlux employs drone-based calibration, since moving a calibration board through a much larger FSF can be prohibitively difficult. In contrast, we introduce a new board-based calibration procedure that remains effective even for large FSF. This eliminates the need for drone-based experiments and significantly simplifies the data collection process.

\subsubsection{Large FSF Board-based Calibration.}
\label{subsubsec:real_calib}

Obtaining high-quality board-based calibration data requires moving the calibration board throughout the camera's FOV while ensuring that the pattern remains visible, in focus, and observed under a wide range of rotations. This becomes difficult in practice for large FSF settings, which typically occur when the camera is zoomed out and FD is large. The board must be placed far from the camera to stay in focus; consequently, it must also have a large pattern to remain visible to the camera from that distance. However, large boards are cumbersome to move, making it difficult to sufficiently excite rotational axes and cover the full FSF.

Rather than moving a large board around, we instead move our camera around a static large board. This still allows the board to assume a variety of poses and angles relative to the camera---note that there is no requirement for the camera to stay static for calibration. Hence, by moving and rotating the camera, we can cover the full FSF and excite all axes of rotation with the board, which is much easier than moving the board for large FSF settings.

For our large FSF experiments, the calibration target we use is an array of AprilTags~\cite{apriltag} projected onto a $5.45~\text{m} \times 3.06~\text{m}$ rigid screen. The screen is installed in a multi-story lecture hall with a balcony. We record footage from four camera positions: left and right on the ground floor, and left and right on the balcony. At each position, we vary the camera rotation, effectively moving the board through the camera FOV. This simultaneously covers the full FSF while exciting rotational axes. See our supplement for additional details on the lenses and calibration experiments used to construct the LUTs.

\begin{table}[t]
  \centering
  \tiny
  \caption{Scene category video counts for InFlux and InFlux++ Real.}
  \label{tab:scene_diversity}
  \setlength{\tabcolsep}{6pt}
  \begin{tabular}{l rr}
    \toprule
    \textbf{Scene Category} & \textbf{InFlux} & \textbf{InFlux++ Real} \\
    \midrule
    \textbf{INDOOR} & \textbf{113} & \textbf{123} \\
    \quad Controlled \& Staged & 41 & 14 \\
    \quad Workspace \& Academic & 60 & 53 \\
    \quad Everyday \& Domestic & 0 & 22 \\
    \quad Casual Activity \& Play & 6 & 4 \\
    \quad Tabletop \& Close-up & 6 & 24 \\
    \quad Elevated \& Aerial Vantage & 0 & 6 \\
    \midrule
    \textbf{OUTDOOR} & \textbf{273} & \textbf{211} \\
    \quad Campus \& Architecture & 95 & 20 \\
    \quad Nature \& Parks & 53 & 3 \\
    \quad Urban & 0 & 20 \\
    \quad Suburban \& Street & 91 & 64 \\
    \quad Monuments \& Landmarks & 9 & 0 \\
    \quad Nighttime & 25 & 23 \\
    \quad Sports \& Recreation & 0 & 58 \\
    \quad Elevated \& Aerial Vantage & 0 & 3 \\
    \quad Moving Vehicle & 0 & 20 \\
    \midrule
    \textbf{Total} & \textbf{386} & \textbf{334} \\
    \bottomrule
  \end{tabular}
\end{table}

\subsubsection{Face and License Plate Blurring.}
\label{subsubsec:real_blur}

To protect the privacy of subjects in our real-world benchmark, we apply modified versions of~\cite{retinaface} and~\cite{egoblur} to every frame to detect and blur faces and license plates. See our supplement for details.

\subsubsection{Data Diversity.}
\label{subsubsec:real_diversity}

We capture 514K+ images across 334 videos. The resolution of our footage is $3424 \times 2202$, filmed at 23.976 FPS.
\\
\\
\textbf{Scene Diversity.}
Compared to InFlux, InFlux++ Real exhibits greater diversity in both scene types and activities. InFlux outdoor footage is largely shot in campus environments, suburban streets, and natural settings. In contrast, InFlux++ Real includes a broader range of urban environments and scenes involving active human motion, such as doing chores and sports. It also introduces new types of captures such as moving-vehicle footage and elevated viewpoints. See \cref{tab:scene_diversity} for a comparison of video category statistics and \cref{fig:real_diversity} for a gallery.
\\
\\
\textbf{Camera Motion Diversity.}
In InFlux, camera motion is mostly rotational. In contrast, InFlux++ Real contains substantially more translational motion, with many sequences captured while the camera is handheld, mounted on a wearable rig, or carried on a moving vehicle. This produces more parallax and viewpoint changes; see our supplement for a qualitative comparison of parallax in InFlux++ Real and InFlux. Because InFlux++ Real videos have extended movement through scenes, its average video length is also longer.
\\
\\
\textbf{Lens and Intrinsics Diversity.} We use the same lens types as~\cite{influx}, but different physical lenses of the same type can have different intrinsics LUTs. One of our lenses has a large principal point offset, expanding the range of $c_x$ and $c_y$ values in InFlux++ Real compared to InFlux. See our supplement for details.

\section{Experiments}
\subsection{Evaluation Metrics and Procedure}

To evaluate predicted intrinsics against ground truth, we report recall at several thresholds for CFL and principal point percent error. We also report EPE recall at several thresholds, but modify its computation~\cite{influx} to ensure metric reliability.

First, to ensure EPE is computed only over physically visible points, we revise EPE's 3D point visibility filter. The original protocol~\cite{influx} keeps any candidate 3D point whose 2D projection via ground truth intrinsics and distortion falls within the image. But under extreme Brown-Conrady~\cite{brown} distortion, the polynomial radial model can become non-monotonic and incorrectly map far-off-axis points back into the image, even though a real lens would not observe these points. We fix this by retaining only points on the main monotonic branch of the radial mapping, cutting off the mapping at a radius $r_*$. See our supplement for details.

Second, to avoid evaluating EPE recall on InFlux++ Real or InFlux frames whose ground truth intrinsics may be unreliable due to LUT interpolation, we define LUT-reliable EPE recall. For each LUT experiment, we estimate LUT interpolation reliability by performing leave-one-out (LOO) validation~\cite{influx} and comparing the interpolated and measured intrinsics via EPE recall@$T$~pixels (px). When computing LUT-reliable EPE recall@$T$~px, we include only frames whose enclosing LUT region has all vertices with LOO EPE recall@$T$~px $\geq 0.95$. Unless otherwise specified, all EPE metrics reported in this paper use the LUT-reliable formulation. See our supplement for details.

\subsection{Baseline Evaluation using InFlux++ Real and InFlux}

\begin{figure}[t]
    \centering
        \includegraphics[width=0.24\linewidth]{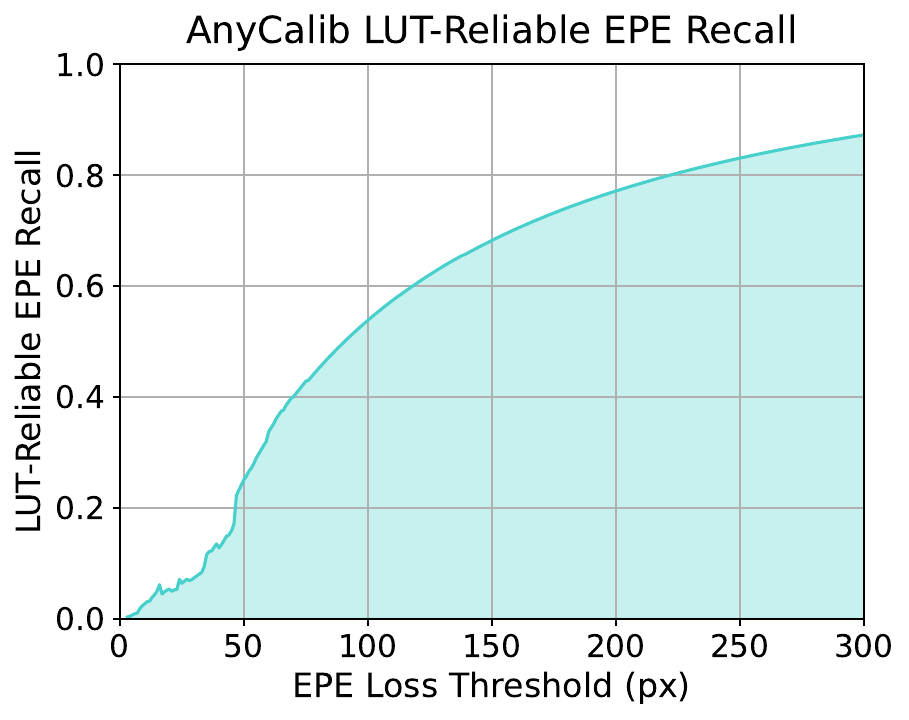}
        \includegraphics[width=0.24\linewidth]{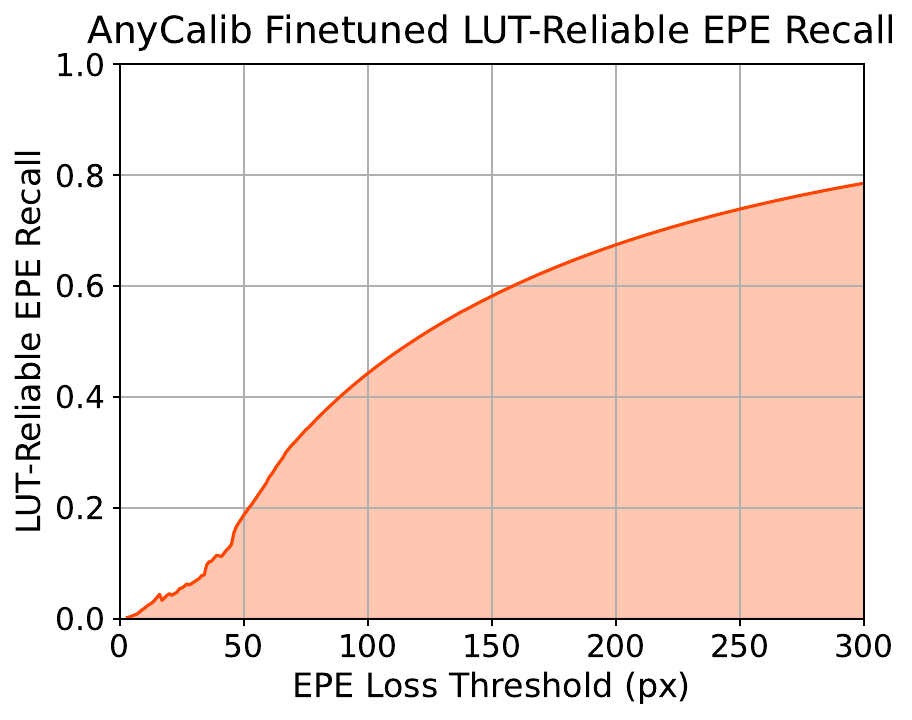}
        \includegraphics[width=0.24\linewidth]{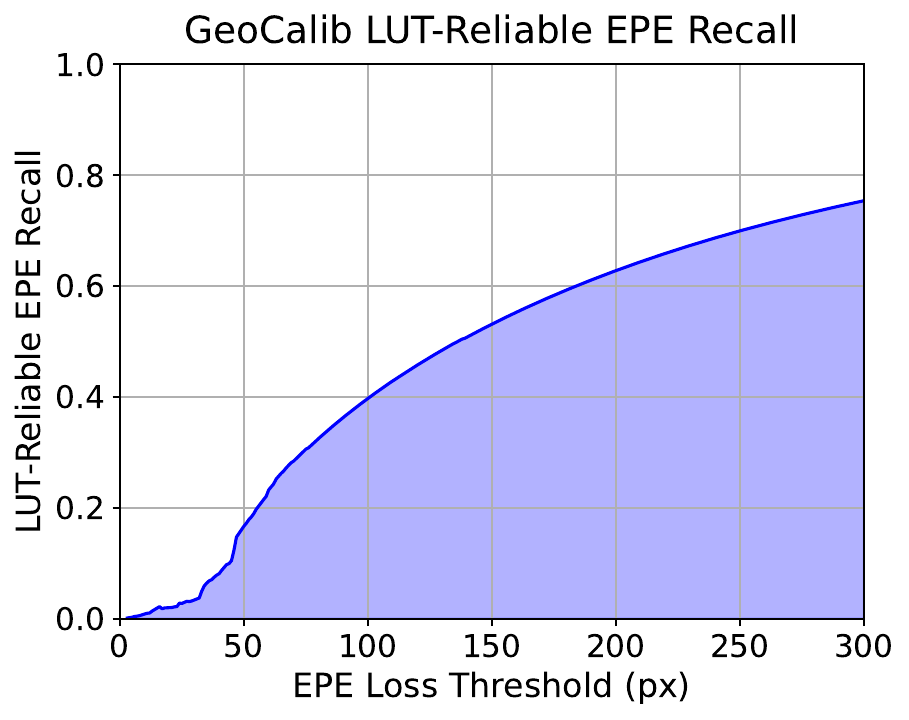}
        \includegraphics[width=0.24\linewidth]{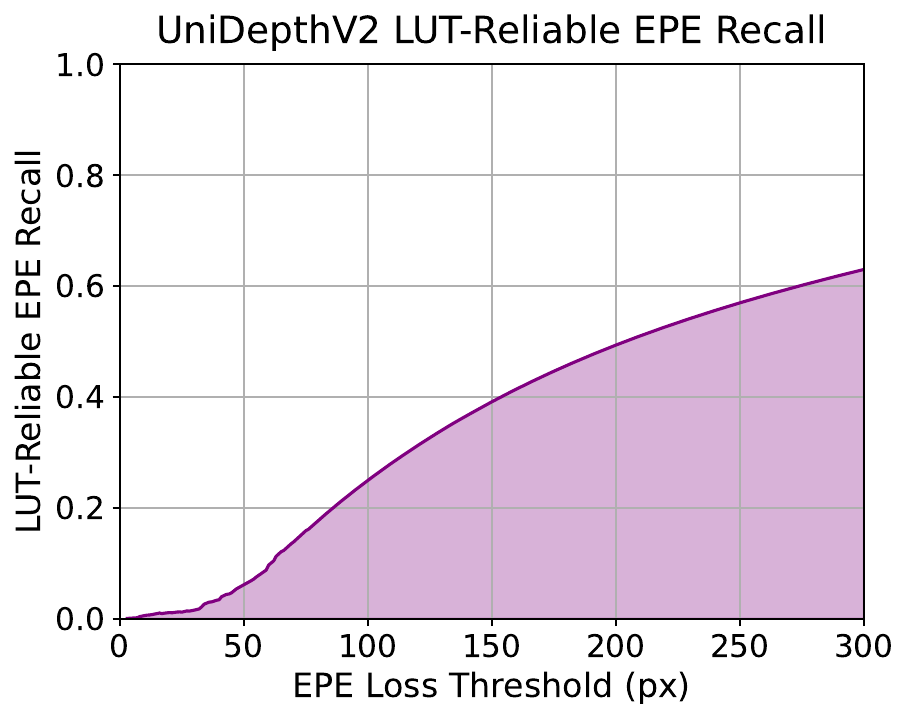}
        \\
        \includegraphics[width=0.24\linewidth]{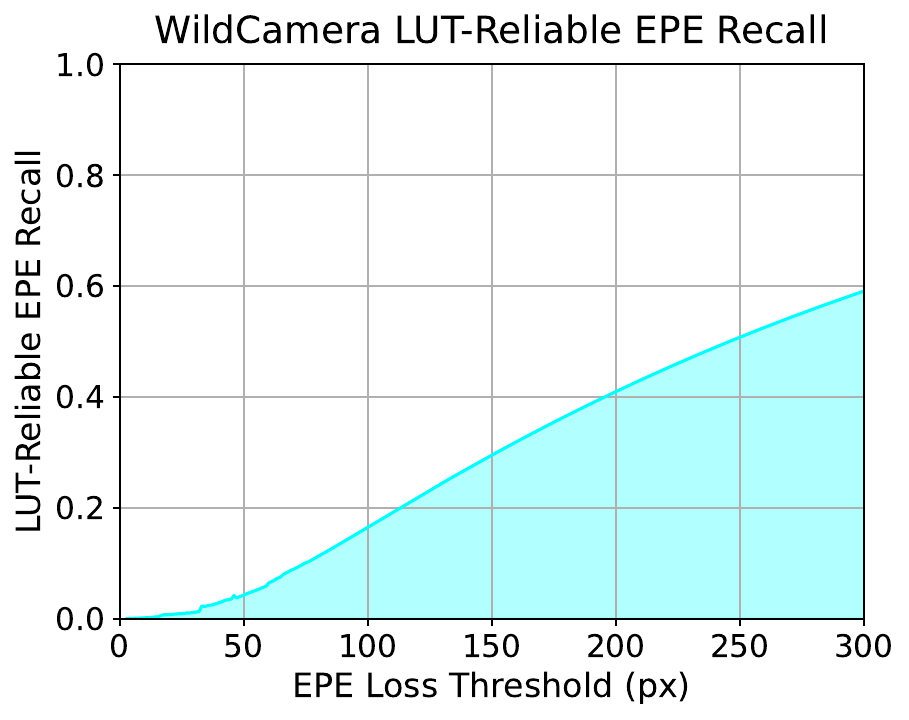}
        \includegraphics[width=0.24\linewidth]{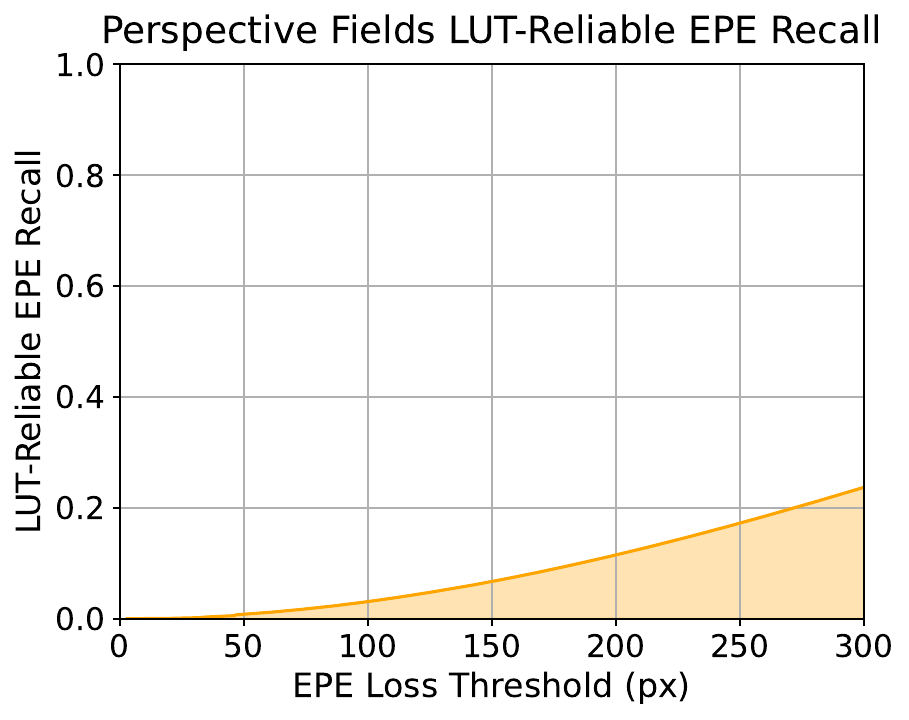}
        \includegraphics[width=0.24\linewidth]{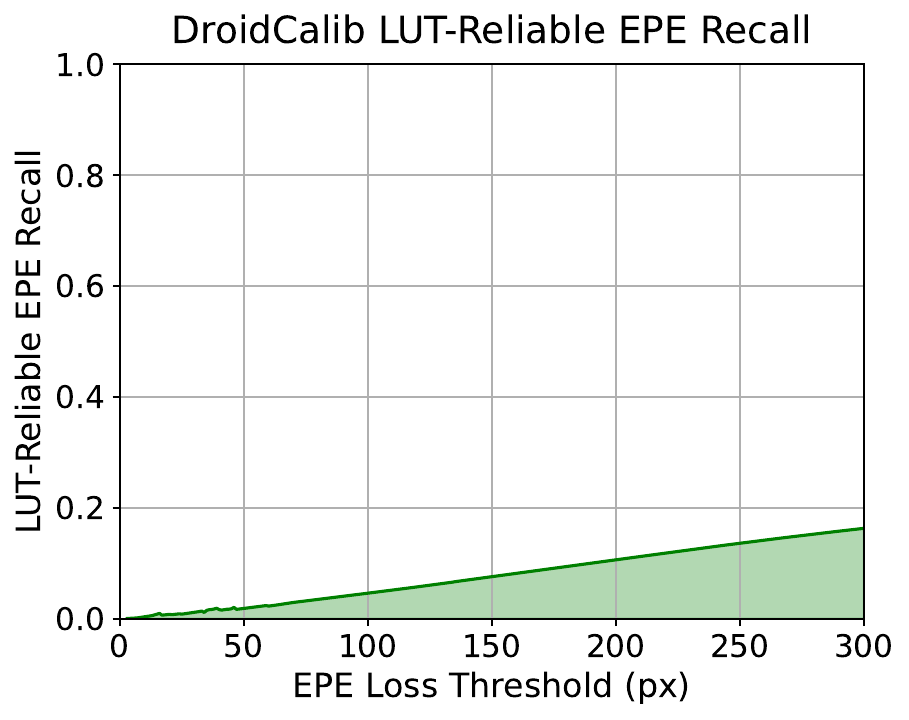}
        \includegraphics[width=0.24\linewidth]{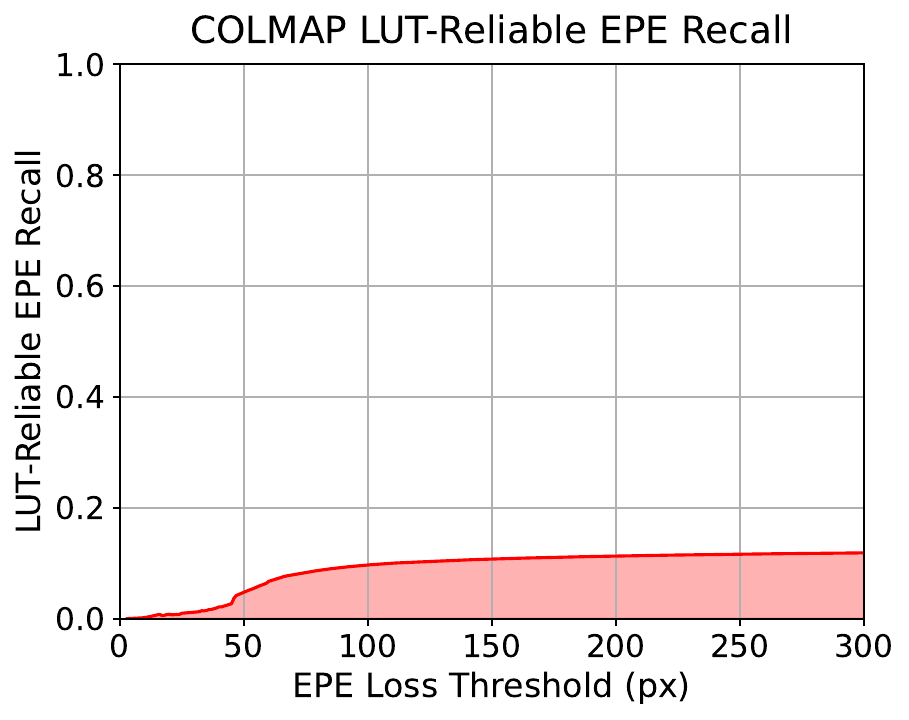}
    \caption{LUT-reliable EPE recall plots for baseline methods, evaluated on the test splits of InFlux++ Real and InFlux for every integer threshold between 1~px and 300~px. AnyCalib performs the best, but still has low EPE recall for low pixel thresholds.}
    \label{fig:epe_thresholds}
\end{figure}

\begin{table*}[t]
    \caption{Baseline intrinsics prediction results on the test splits of InFlux and InFlux++ Real. We report recall at multiple thresholds for $f_x$, $f_y$, $c_x$, and $c_y$ percent error, as well as LUT-reliable EPE recall at multiple pixel thresholds. All recalls are reported as percentages; higher is better. Best is \textbf{bolded}, second is \underline{underlined}.}
    \label{tab:baseline_methods}
    \centering
    \scriptsize
    \setlength{\tabcolsep}{2.0pt}
    \renewcommand{\arraystretch}{0.95}
    \adjustbox{max width=\textwidth}{%
        \begin{tabular}{@{}l ccc ccc ccc ccc *{3}{E}@{}}
            \toprule
            & \multicolumn{3}{c}{$\mathbf{f_x}$ \textbf{Recall (\%)}}
            & \multicolumn{3}{c}{$\mathbf{f_y}$ \textbf{Recall (\%)}}
            & \multicolumn{3}{c}{$\mathbf{c_x}$ \textbf{Recall (\%)}}
            & \multicolumn{3}{c}{$\mathbf{c_y}$ \textbf{Recall (\%)}}
            & \epegroup{\textbf{LUT-Reliable EPE Recall (\%)}} \\
            \cmidrule(lr){2-4}
            \cmidrule(lr){5-7}
            \cmidrule(lr){8-10}
            \cmidrule(lr){11-13}
            \cmidrule(lr){14-16}
            \textbf{Method}
            & \textbf{@1\%} & \textbf{@10\%} & \textbf{@20\%}
            & \textbf{@1\%} & \textbf{@10\%} & \textbf{@20\%}
            & \textbf{@0.5\%} & \textbf{@1\%} & \textbf{@2\%}
            & \textbf{@0.5\%} & \textbf{@1\%} & \textbf{@2\%}
            & \textbf{@10~px} & \textbf{@50~px} & \textbf{@300~px} \\
            \midrule
            AnyCalib~\cite{anycalib}
            & \underline{3.00} & \underline{25.2} & \underline{41.3}
            & \underline{2.96} & \underline{25.2} & \underline{41.3}
            & \underline{41.6} & \underline{76.5} & \underline{98.1}
            & \underline{24.1} & \underline{37.4} & \textbf{46.2}
            & \textbf{2.70} & \textbf{25.1} & \textbf{87.3} \\

            AnyCalib (finetuned)
            & \textbf{3.68} & \textbf{31.2} & \textbf{50.4}
            & \textbf{3.65} & \textbf{31.3} & \textbf{50.5}
            & 33.0 & 67.2 & 95.4
            & 20.4 & 33.4 & 44.1
            & \underline{1.93} & \underline{18.7} & \underline{78.6}\\
            
            \midrule

            GeoCalib~\cite{geocalib}
            & 2.17 & 19.7 & 33.5
            & 2.15 & 19.6 & 33.4
            & \textbf{49.5} & \textbf{82.3} & \textbf{99.4}
            & \textbf{38.3} & \textbf{40.9} & \underline{45.0}
            & 8.53e-1 & 16.7 & 75.4 \\

            UniDepthV2~\cite{unidepthv2}
            & 1.41 & 14.0 & 26.6
            & 1.31 & 13.1 & 25.7
            & 21.8 & 45.4 & 69.9
            & 3.87 & 9.14 & 25.5
            & 5.68e-1 & 6.14 & 63.0 \\

            WildCamera~\cite{wildcamera}
            & 1.29 & 12.9 & 26.6
            & 1.28 & 13.0 & 26.5
            & 4.27 & 8.91 & 20.7
            & 5.38 & 10.8 & 21.3
            & 1.70e-1 & 4.26 & 59.1 \\

            Perspective Fields~\cite{perspectivefields}
            & 7.32e-1 & 7.65 & 16.7
            & 7.07e-1 & 7.53 & 16.5
            & 1.48 & 2.97 & 5.93
            & 1.55 & 3.09 & 6.20
            & 1.29e-1 & 8.06e-1 & 23.7 \\

            DroidCalib~\cite{droidcalib}
            & 7.11e-1 & 5.99 & 11.8
            & 5.66e-1 & 5.95 & 12.0
            & 2.92 & 6.07 & 14.4
            & 2.54 & 4.45 & 8.48
            & 3.33e-1 & 1.86 & 16.3 \\

            COLMAP~\cite{colmap}
            & 1.49 & 6.34 & 7.22
            & 1.56 & 6.36 & 7.23
            & 3.14 & 7.16 & 8.87
            & 1.78 & 1.96 & 2.31
            & 2.25e-1 & 4.78 & 11.9 \\
            
            \bottomrule
        \end{tabular}%
    }
\end{table*}

We evaluate seven baseline intrinsics prediction methods on the test splits of InFlux++ Real and InFlux: AnyCalib~\cite{anycalib}, COLMAP~\cite{colmap}, DroidCalib~\cite{droidcalib}, GeoCalib~\cite{geocalib}, Perspective Fields~\cite{perspectivefields}, UniDepthV2~\cite{unidepthv2}, and WildCamera~\cite{wildcamera}. See our supplement for details on the InFlux++ Real test split. For all baselines except AnyCalib (AC), we use the model settings specified in~\cite{influx}. For AC, we use the \texttt{anycalib\_gen.pt} weights available on Hugging Face.  We report recall@\{1, 10, 20\}\% for $f_x$ and $f_y$ percent errors, recall@\{0.5, 1, 2\}\% for $c_x$ and $c_y$ percent errors, and recall@\{10, 50, 300\}~px for LUT-reliable EPE. See \cref{tab:baseline_methods} for a table of baseline performance and \cref{fig:epe_thresholds} for a plot of LUT-reliable EPE recall at additional thresholds. We additionally provide a breakdown of AC performance on InFlux++ Real versus InFlux in \cref{tab:finetune_results}.

Overall, the results show that dynamic intrinsics prediction remains challenging. AC performs best, yet only achieves 25.2\% for $f_x$ recall@10\% and 25.1\% for EPE recall@50~px. AC achieves higher $f_x$, $f_y$, and EPE recall on InFlux++ Real than on InFlux, but worse $c_x$ and $c_y$ recall. This may be because InFlux++ Real has proportionally more indoor scenes than InFlux, which better matches AC's indoor-heavy training distribution. InFlux++ Real contains more off-center principal point frames, which could make accurate $c_x$ and $c_y$ prediction harder.

\subsection{AnyCalib Finetuning using InFlux++ Synth}

We explore the idea of using synthetic supervision for dynamic intrinsics prediction by finetuning AC with a subset of InFlux++ Synth. We perform finetuning for 15 epochs and take the checkpoint with the lowest CFL percent error on the validation split of InFlux++ Real. See our supplement for details.

As shown in \cref{tab:finetune_results}, finetuning AC on InFlux++ Synth improves recall for $f_x$ and $f_y$ error at all thresholds. However, it worsens recall for $c_x$, $c_y$, and EPE, with the exception of EPE recall@10~px for InFlux++ Real and EPE recall@300~px for InFlux. One possible explanation is that AC's FOV-field training loss is not directly aligned with EPE, especially near image boundaries where lens distortion is most pronounced. This may limit the model's ability to fit accurate distortion parameters, which strongly affect EPE; see our supplement for details. Nevertheless, these results suggest that InFlux++ Synth is a promising source of supervision for dynamic intrinsics prediction models.

\begin{table*}[t]
    \caption{Evaluation of AnyCalib before and after finetuning on InFlux++ Synth. Finetuning improves CFL estimation. Best is \textbf{bolded} within each evaluation set.}
    \label{tab:finetune_results}
    \centering
    \scriptsize
    \setlength{\tabcolsep}{2.0pt}
    \renewcommand{\arraystretch}{0.95}
    \adjustbox{max width=\textwidth}{%
        \begin{tabular}{@{}c l ccc ccc ccc ccc *{3}{E}@{}}
            \toprule
            & & \multicolumn{3}{c}{$\mathbf{f_x}$ \textbf{Recall (\%)}}
            & \multicolumn{3}{c}{$\mathbf{f_y}$ \textbf{Recall (\%)}}
            & \multicolumn{3}{c}{$\mathbf{c_x}$ \textbf{Recall (\%)}}
            & \multicolumn{3}{c}{$\mathbf{c_y}$ \textbf{Recall (\%)}}
            & \epegroup{\textbf{LUT-Reliable EPE Recall (\%)}} \\
            \cmidrule(lr){3-5}
            \cmidrule(lr){6-8}
            \cmidrule(lr){9-11}
            \cmidrule(lr){12-14}
            \cmidrule(lr){15-17}
            \textbf{Eval. Set} & \textbf{Method}
            & \textbf{@1\%} & \textbf{@10\%} & \textbf{@20\%}
            & \textbf{@1\%} & \textbf{@10\%} & \textbf{@20\%}
            & \textbf{@0.5\%} & \textbf{@1\%} & \textbf{@2\%}
            & \textbf{@0.5\%} & \textbf{@1\%} & \textbf{@2\%}
            & \textbf{@10~px} & \textbf{@50~px} & \textbf{@300~px} \\
            \midrule
            \multirow{2}{*}{InFlux}
            & AnyCalib~\cite{anycalib}
            & 1.30 & 11.5 & 20.8
            & 1.38 & 11.7 & 21.0
            & \textbf{73.5} & \textbf{94.4} & \textbf{99.6}
            & \textbf{51.3} & \textbf{81.7} & \textbf{99.2}
            & \textbf{2.95} & \textbf{34.1} & 80.1 \\
            & AnyCalib (finetuned)
            & \textbf{2.21} & \textbf{17.5} & \textbf{34.1}
            & \textbf{2.23} & \textbf{18.7} & \textbf{34.9}
            & 65.5 & 90.0 & 98.5
            & 47.4 & 76.1 & 98.3
            & 2.09 & 27.9 & \textbf{82.7} \\

            \midrule
            \multirow{2}{*}{InFlux++ Real}
            & AnyCalib~\cite{anycalib}
            & 3.36 & 28.2 & 45.7
            & 3.31 & 28.1 & 45.7
            & \textbf{34.7} & \textbf{72.6} & \textbf{97.8}
            & \textbf{18.2} & \textbf{27.8} & \textbf{34.7}
            & 6.40e-3 & \textbf{24.3} & \textbf{87.9} \\
            & AnyCalib (finetuned)
            & \textbf{4.00} & \textbf{34.2} & \textbf{54.0}
            & \textbf{3.96} & \textbf{34.0} & \textbf{53.9}
            & 25.9 & 62.3 & 94.7
            & 14.5 & 24.2 & 32.3
            & \textbf{1.66e-1} & 17.9 & 78.2 \\

            \midrule
            \multirow{2}{*}{Combined}
            & AnyCalib~\cite{anycalib}
            & 3.00 & 25.2 & 41.3
            & 2.96 & 25.2 & 41.3
            & \textbf{41.6} & \textbf{76.5} & \textbf{98.1}
            & \textbf{24.1} & \textbf{37.4} & \textbf{46.2}
            & \textbf{2.70} & \textbf{25.1} & \textbf{87.3} \\
            & AnyCalib (finetuned)
            & \textbf{3.68} & \textbf{31.2} & \textbf{50.4}
            & \textbf{3.65} & \textbf{31.3} & \textbf{50.5}
            & 33.0 & 67.2 & 95.4
            & 20.4 & 33.4 & 44.1
            & 1.93 & 18.7 & 78.6 \\
            \bottomrule
        \end{tabular}%
    }
\end{table*}

\section{Conclusion}
We present InFlux++, a data suite for dynamic intrinsics prediction. InFlux++ Synth provides synthetic videos with accurate and dynamic intrinsics, diverse scenes and materials, dynamic objects, and realistic optical effects such as defocus blur. InFlux++ Real expands upon the existing real-world benchmark InFlux with greater scene and camera motion diversity. Current methods struggle to predict accurate intrinsics on InFlux++ Real and InFlux. However, finetuning with InFlux++ Synth data is a promising direction for improving intrinsics prediction. Through InFlux++, we aim to support and encourage future progress in dynamic intrinsics prediction.

\section*{Acknowledgements}
This work was partially supported by the National Science Foundation. Erich Liang was supported in part by the NSF Graduate Research Fellowship Program under Grant No. 2146752. We thank our friends and colleagues at Princeton University for their help with filming the benchmark.

\bibliographystyle{splncs04}
\bibliography{main}

\newpage
\appendix
\input{supplement}

\end{document}

%% file: supplement.tex
\renewcommand{\thefigure}{\Alph{figure}}
\setcounter{figure}{0}

\renewcommand{\thetable}{\Alph{table}}
\setcounter{table}{0}
\setcounter{tocdepth}{3}

\section*{Appendix}

\section{Additional Details on InFlux++ Synth: Camera Pose, Depth, Surface Normals, and Other Annotations}

For each video in InFlux++ Synth, we provide per-frame RGB, ground truth intrinsics, and camera pose. For a subset of the videos, we also provide per-frame depth and surface normals. See \cref{tab:influx_synth_composition} for statistics on annotation coverage.
\\
\\
\textbf{Camera intrinsics and pose.} For each frame, InFlux++ Synth provides the intrinsics matrix $K$, the camera-to-world matrix $T$, along with additional camera metadata such as image resolution, lens to object distance (LTO), lens focal length (LFL), and camera focal length (CFL). We report LFL and CFL in millimeters, and we report LTO in meters. Because image distortion is applied by our data loader as data augmentation, no distortion coefficients are provided.

\begin{figure}[t]
    \centering
    \includegraphics[width=\linewidth]{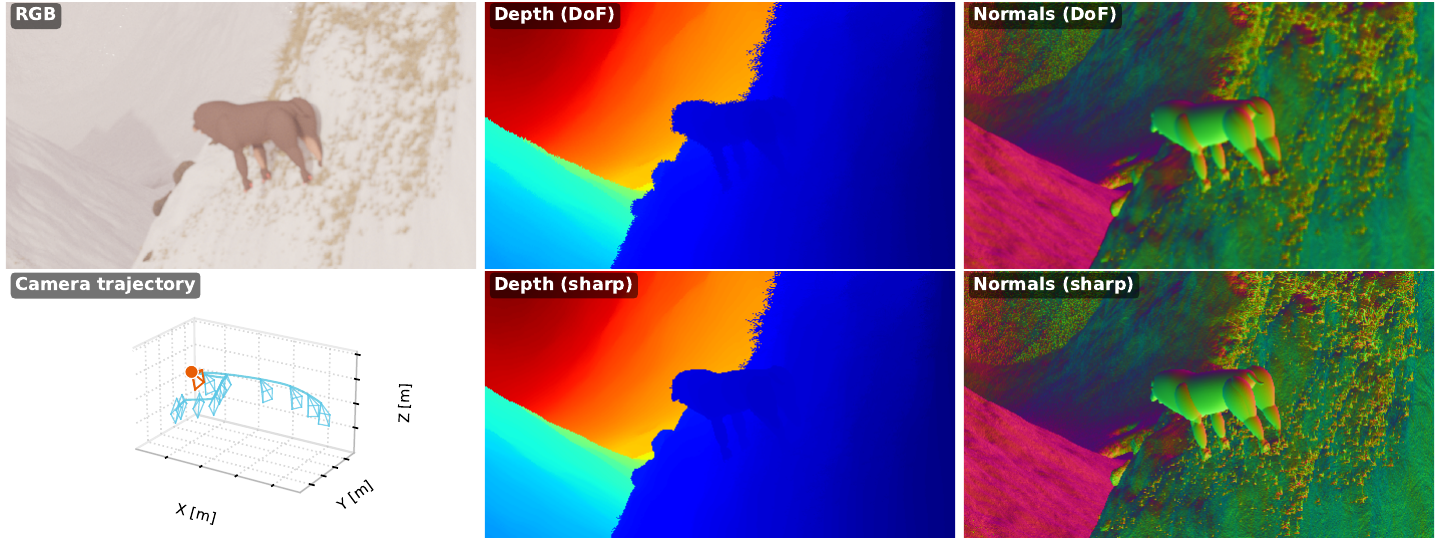} 
    \caption{Visualization of ground truth modalities for a frame from InFlux++ Synth. \textit{(Left Column)}: the top image shows the RGB image rendered with DoF and the bottom image shows the camera's trajectory in the scene. \textit{(Center Column)}: the top image shows the depth map rendered with DoF, while the bottom image shows the depth map rendered with a pinhole camera. \textit{(Right Column)}: the top image shows the surface normal map rendered with DoF, while the bottom image shows the surface normal map rendered with a pinhole camera. Differences between DoF and sharp annotations are most visible in vegetation and at occlusion boundaries such as the foliage along the cliffs and around the mammal's head.}
    \label{fig:gt_modality_gallery}
\end{figure}

We follow Infinigen's~\cite{infinigen} coordinate conventions. Specifically for $T$, the camera coordinate system follows $+X$ as right, $+Y$ as down, and $+Z$ as forward. Pixel coordinates use center-at-half-integer indexing. For example, the center of an image of size $W \times H$ is $(W/2,H/2)$.
\\
\\
\textbf{Depth and surface normals.} For a subset of InFlux++ Synth scenes, we provide per-frame depth and surface normals. Each modality is released in two variants: a defocus-blurred version and a sharp version. The blurred version is rendered with depth of field (DoF) using Blender's thin lens camera model, so each pixel aggregates information over sampled aperture rays contributing to that pixel. The sharp version is rendered without DoF and corresponds to an ideal pinhole camera model. Depth is reported in meters, and surface normals are reported as unit vectors in camera space, but with $+X$ as right, $+Y$ as up, and $+Z$ as backward. See \cref{fig:gt_modality_gallery} for a visualization of the ground truth annotations for a frame of a nature scene.

\section{Additional Details on InFlux++ Synth: Camera Motion}

We build on the Rapidly-Exploring Random Trees (RRT) planner in Infinigen~\cite{infinigen}, but we augment its view validation to favor views with richer geometric structure to produce more visually informative training data for dynamic intrinsics prediction. Because these stricter camera pose requirements increase the chances of trajectory generation failure, we modify the planner's recovery logic to keep valid motion whenever possible.
\\
\\
\textbf{Original Infinigen RRT Planner.} Infinigen constructs camera trajectories incrementally from a sequence of RRT segments. Starting from an accepted camera pose, the planner creates an RRT segment by sampling a target position and querying RRT to obtain a sequence of positional waypoints towards it. The planner then assigns camera orientations to waypoints sequentially: for each waypoint, it samples an orientation, connects the resulting pose to the previous one via Bézier interpolation, and validates the resulting camera motion. After all waypoints have been assigned an orientation, the planner creates a new RRT segment and repeats this process until a desired number of frames is reached. In our configuration, Infinigen oversamples initial camera views, attempts RRT trajectory generation from the retained views, scores each successful trajectory, and retains the highest scoring proposal as the final trajectory.

\begin{table}[t]
    \centering
    \small
    \caption{Composition of the InFlux++ Synth dataset by scene type and annotation coverage. All videos are 240 frames and have per-frame RGB, intrinsics, and camera pose. A subset of videos additionally has per-frame depth, sharp depth, surface normals, and sharp surface normals.}
    \label{tab:influx_synth_composition}
    \begin{tabular}{lrrrrrr}
        \toprule
        & \multicolumn{2}{c}{Indoor} &
          \multicolumn{2}{c}{Nature} &
          \multicolumn{2}{c}{Total} \\
        \cmidrule(lr){2-3}
        \cmidrule(lr){4-5}
        \cmidrule(lr){6-7}
        Subset & Videos & Frames & Videos & Frames & Videos & Frames \\
        \midrule
        Without depth and surface normals & 515 & 123{,}600 & 508 & 121{,}920 & 1{,}023 & 245{,}520 \\
        With depth and surface normals & 525 & 126{,}000 & 293 & 70{,}320 & 818 & 196{,}320 \\
        \midrule
        Total & 1{,}040 & 249{,}600 & 801 & 192{,}240 & 1{,}841 & 441{,}840 \\
        \bottomrule
    \end{tabular}
\end{table}

The planner is not always successful at assigning an acceptable camera orientation to each waypoint, because Infinigen rejects views containing excessive sky or nearby geometry. In our configuration, the proximity check rejects a view when more than $75\%$ of the sampled rays intersect geometry closer than a scene-specific distance threshold. The original implementation uses a fixed retry budget when attempting to assign a pose to the next waypoint; this budget resets after a waypoint is successfully accepted. If orientation assignment fails at the first waypoint of an RRT segment, the next retry generates a new RRT segment from the same starting pose and attempts orientation assignment for the first waypoint of that new segment. If orientation assignment fails at a later waypoint, the planner keeps the same waypoint position and samples a new orientation. If the retry budget is exhausted before the next waypoint is accepted, the entire trajectory proposal is discarded.
\\
\\
\textbf{Dominant Surface Normal View Validation.} In addition to the sky and proximity checks that the original planner performs, we add a new criterion in our planner that rejects camera orientations in which a single spatially coherent surface dominates the view. This is useful for producing better dynamic intrinsics training data, as it helps avoid admitting views dominated by a single broad surface with little geometric variation such as blank walls, floors, and ceilings in indoor scenes and featureless terrain in nature scenes.

To achieve this, we screen candidate orientations at a waypoint by sparsely sampling the resulting views' surface normals and depths to estimate whether they are dominated by one geometrically uniform surface. Specifically, for each orientation, we cast at least 400 rays arranged on a regular grid over the image plane. For every ray that intersects non-sky geometry, we record the visible surface normal $\mathbf{n}_i$ and its depth $d_i$ along the camera optical axis. We use singular value decomposition over the sampled surface normals to estimate the dominant surface normal axis $\mathbf{n}_{\mathrm{dom}}$, and identify sampled locations whose normals are aligned with this dominant axis:
\begin{equation*}
    \left|
    \mathbf{n}_i^\top \mathbf{n}_{\mathrm{dom}}
    \right| > 0.9847.
\end{equation*}
This corresponds to an angular tolerance of approximately $10^\circ$.

Normal alignment alone does not imply that two samples belong to the same physical surface. For example, a tabletop and the floor may have parallel surface normals but are at two different depths. To account for this, we use a depth-based heuristic to find the largest connected component of dominant axis-aligned samples that may lie on the same surface. Two horizontally or vertically adjacent samples $i$ and $j$ are connected only if
\begin{equation*}
    \left| d_i - d_j \right| \leq 0.2\,\mathrm{m}.
\end{equation*}
We denote the largest resulting component by $\mathcal{C}_{\max}$ and compute the fraction of sampled locations outside this component as
\begin{equation*}
    q_{\mathrm{outside}} = 1 - \frac{|\mathcal{C}_{\max}|}{N},
\end{equation*}
where $N$ is the total number of image-plane sample locations, rather than only the number of dominant axis-aligned samples. We require $q_{\mathrm{outside}} \geq 0.35$ for indoor scenes and $q_{\mathrm{outside}} \geq 0.20$ for nature scenes. In other words, the largest normal-aligned surface component may occupy at most $65\%$ of the sampled view for indoor scenes or $80\%$ for nature scenes. The more permissive nature scene threshold accommodates broad terrain regions that may still provide useful geometric structure.
\\
\\
\textbf{Orientation Resampling and Local Replanning.} Our stricter view-content criterion makes Infinigen's original retry strategy inefficient: more candidate orientations are rejected, which increases the chance that one unsuitable waypoint causes the entire trajectory to be discarded. However, failure to find an orientation at one waypoint is often an issue local to a certain part of the scene, and should not invalidate the previously completed RRT segments. We therefore use a staged recovery strategy. First, we allocate a dedicated budget for resampling orientations while keeping the waypoint position fixed. If this search is exhausted, we preserve all previously completed RRT segments and replan only the current RRT segment, rather than restarting the trajectory from scratch.

Before generating any RRT segments, Infinigen proposes a set of candidate initial camera poses from which to begin trajectory generation, but their camera orientations may not pass our view-content checks. For each candidate, we first evaluate the proposed orientation. If it fails validity checks, we keep the position fixed and test up to 29 additional orientations sampled by adding independent zero-mean Gaussian perturbations with a standard deviation of $20^\circ$ to each Euler angle. If none of the 30 orientations passes validation, we discard that initial pose candidate and proceed to the next proposal.

For subsequent waypoints, to find a valid orientation, we similarly evaluate up to 50 candidate orientations. We apply independent zero-mean Gaussian perturbations with a standard deviation of $8^\circ$ per Euler angle. If all orientation attempts fail validity checks, this may indicate that the positional route taken by the current RRT segment passes through a region with few suitable viewing directions. We therefore rewind to the pose at which the current RRT segment began, sample a new target, and generate waypoints for a replacement RRT segment. This preserves all previously completed RRT segments while allowing the replacement segment to find a different route through the scene with more valid orientation options. We allow up to five regenerations of an RRT segment before falling back to a complete trajectory restart.
\\
\\
\textbf{Additional Implementation Details.} We make several smaller implementation changes for robustness and scene-specific correctness. First, we correct the treatment of environment-enclosing geometry during pose validation. Infinigen counts rays that miss scene geometry as sky, but upward rays in nature scenes may instead intersect the enclosing atmosphere mesh. To address this, we classify both ray misses and atmosphere intersections as sky, making the configured sky-fraction threshold effective. In underwater scenes, intersections with liquid surface meshes are treated analogously. Liquid meshes are also excluded from the object set used to determine whether an RRT node lies inside scene geometry, preventing valid underwater camera positions from being rejected by the enclosing water surface.

Second, we enable Infinigen's optional free-space check between consecutive interpolated camera positions. After a new waypoint is keyframed and connected to the previous one, we evaluate the Bézier-interpolated camera centers at intermediate frames. For each consecutive pair of intermediate frames, we raycast from one camera center to the next. If any of these line segments intersects scene geometry, we invoke local RRT segment replanning as described above.

\begin{table}[t]
    \centering
    \small
    \caption{Scene-specific RRT camera-motion parameters for indoor and
    nature scenes.}
    \label{tab:rrt_scene_specific_parameters}
    \begin{tabular}{lcc}
        \toprule
        Parameter & Indoor & Nature \\
        \midrule
        Maximum sky fraction & 0.75 & 0.85 \\
        Near-geometry distance & $1.0\,\mathrm{m}$ & $0.1\,\mathrm{m}$ \\
        Minimum fraction outside dominant surface & 0.35 & 0.20 \\
        \midrule
        RRT step range & $(1,2)\,\mathrm{m}$ & $(2,5)\,\mathrm{m}$ \\
        Camera speed range & $(1,2)\,\mathrm{m/s}$ & $(2,5)\,\mathrm{m/s}$ \\
        Maximum attempts per waypoint transition & 30 & 200 \\
        \midrule
        Valid initial pose candidates targeted & 160 & 80 \\
        Initial views retained for RRT & 16 & 8 \\
        \bottomrule
    \end{tabular}
\end{table}

Finally, we add two small safeguards. First, we reject any waypoint-to-waypoint motion that lasts less than six frames, since very short transitions between waypoints can produce abrupt camera jitter. If this occurs, we trigger local replanning of the current RRT segment. Second, if attempts to generate an RRT path return an empty waypoint sequence five times from the same frame and camera position, we terminate the current trajectory attempt rather than repeatedly searching from the same degenerate state. For nature scenes, we also require initial camera poses to be at least $1.0\,\mathrm{m}$ from coarse scene placeholders, reducing the chance of initializing the camera inside vegetation or other coarse assets.

\cref{tab:rrt_scene_specific_parameters} summarizes additional parameters that differ between indoor and nature scenes. The first block contains view-validation thresholds; the second block contains RRT motion and retry settings; and the final block contains initial-view selection settings used before full RRT trajectory generation.

\section{InFlux++ Synth: Additional Details on Per-Frame Intrinsics Parameterization}

For a physical thin lens, varying the LTO also changes the CFL needed to keep the desired plane in focus. This relationship is governed by the thin lens equation:
\begin{equation*}
    \frac{1}{LFL} = \frac{1}{CFL} + \frac{1}{LTO}
\end{equation*}
where LFL denotes the lens focal length, a fixed physical property of the lens.

Blender's Cycles rendering engine, which is used by Infinigen Indoors~\cite{infinigen2024indoors}, uses the thin lens model to simulate ray bending and defocus blur. However, unlike a physical camera, Cycles allows CFL and LTO to be set independently despite their coupling in the thin lens equation. This is possible because the renderer does not keep LFL fixed. Instead, when the user specifies CFL and LTO, Cycles implicitly adjusts the effective value of LFL so that the thin lens equation is satisfied. As a result, lens breathing does not occur by default in Blender.

\section{InFlux++ Synth: Lens Breathing Visualization}

\begin{figure}[t]
    \centering
    \includegraphics[width=0.24\linewidth]{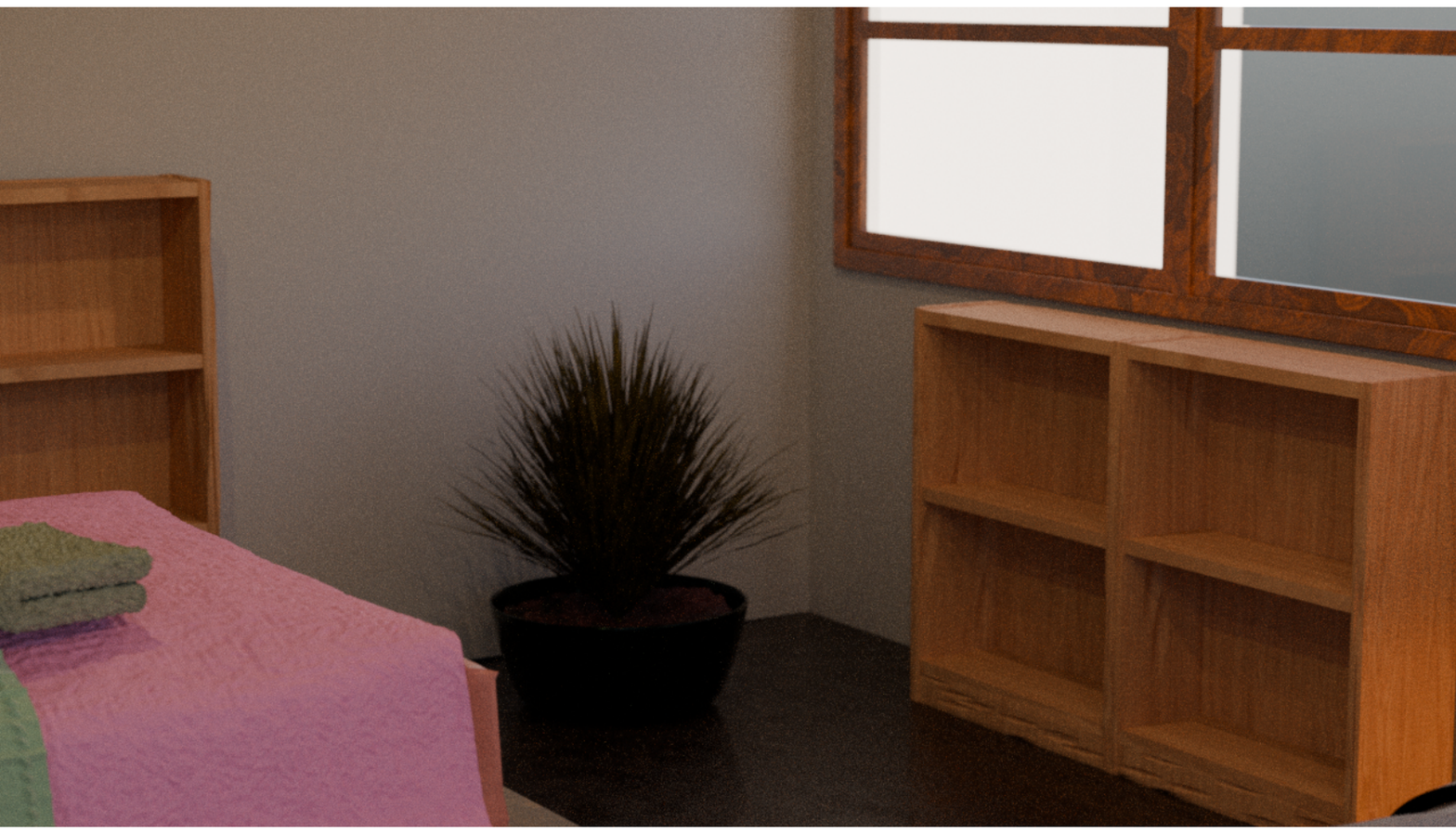}
    \includegraphics[width=0.24\linewidth]{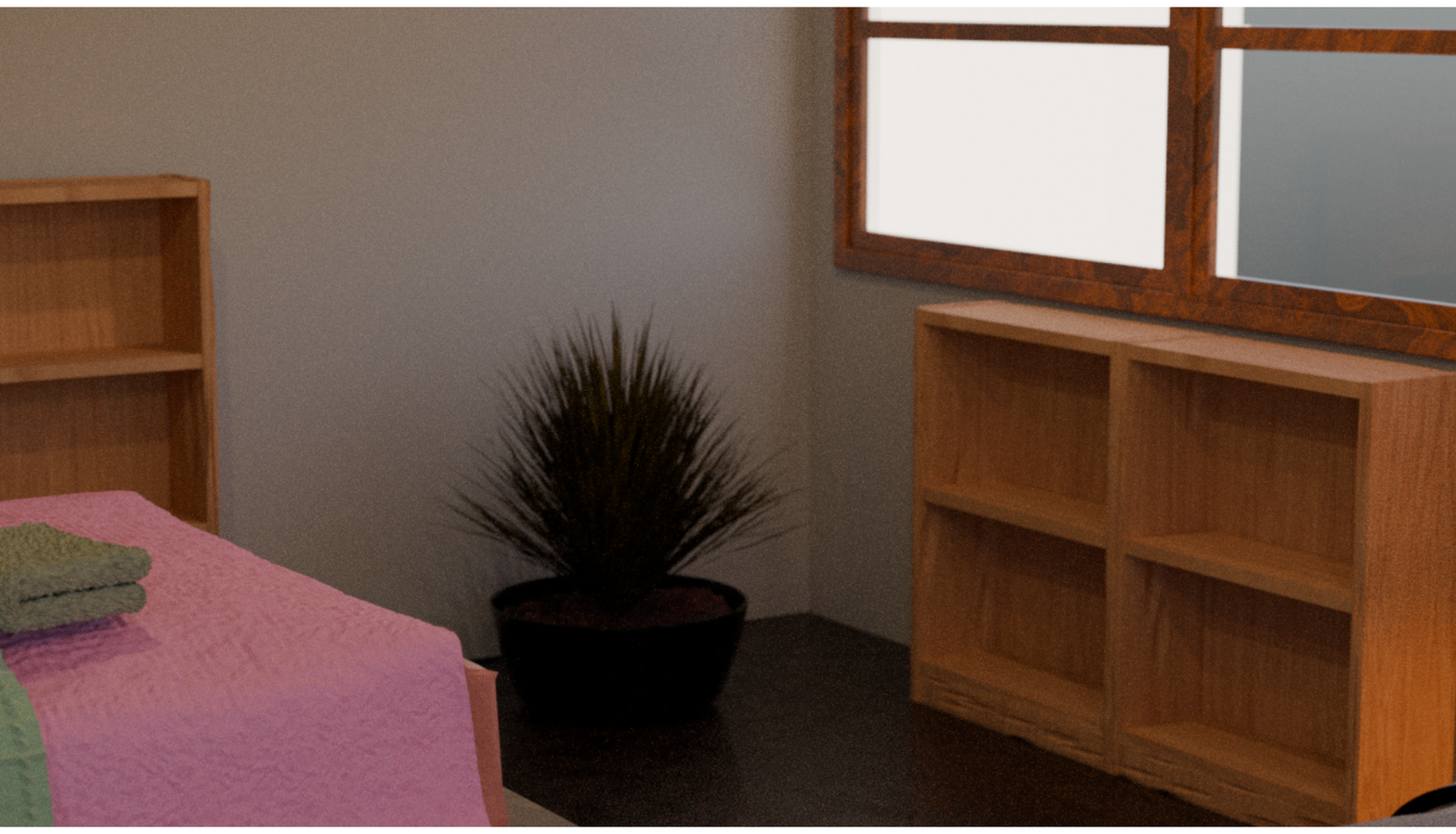}
    \includegraphics[width=0.24\linewidth]{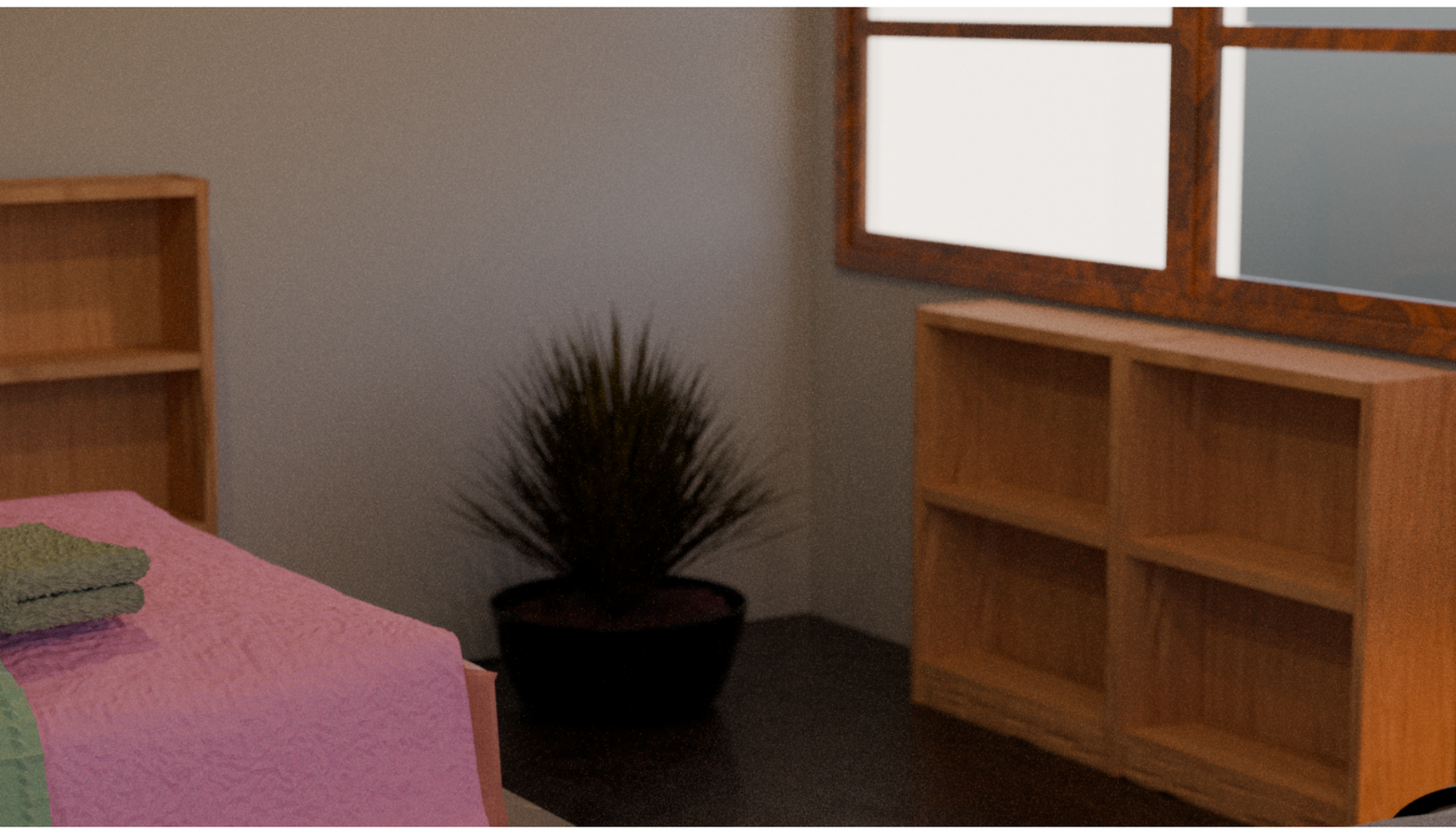}
    \includegraphics[width=0.24\linewidth]{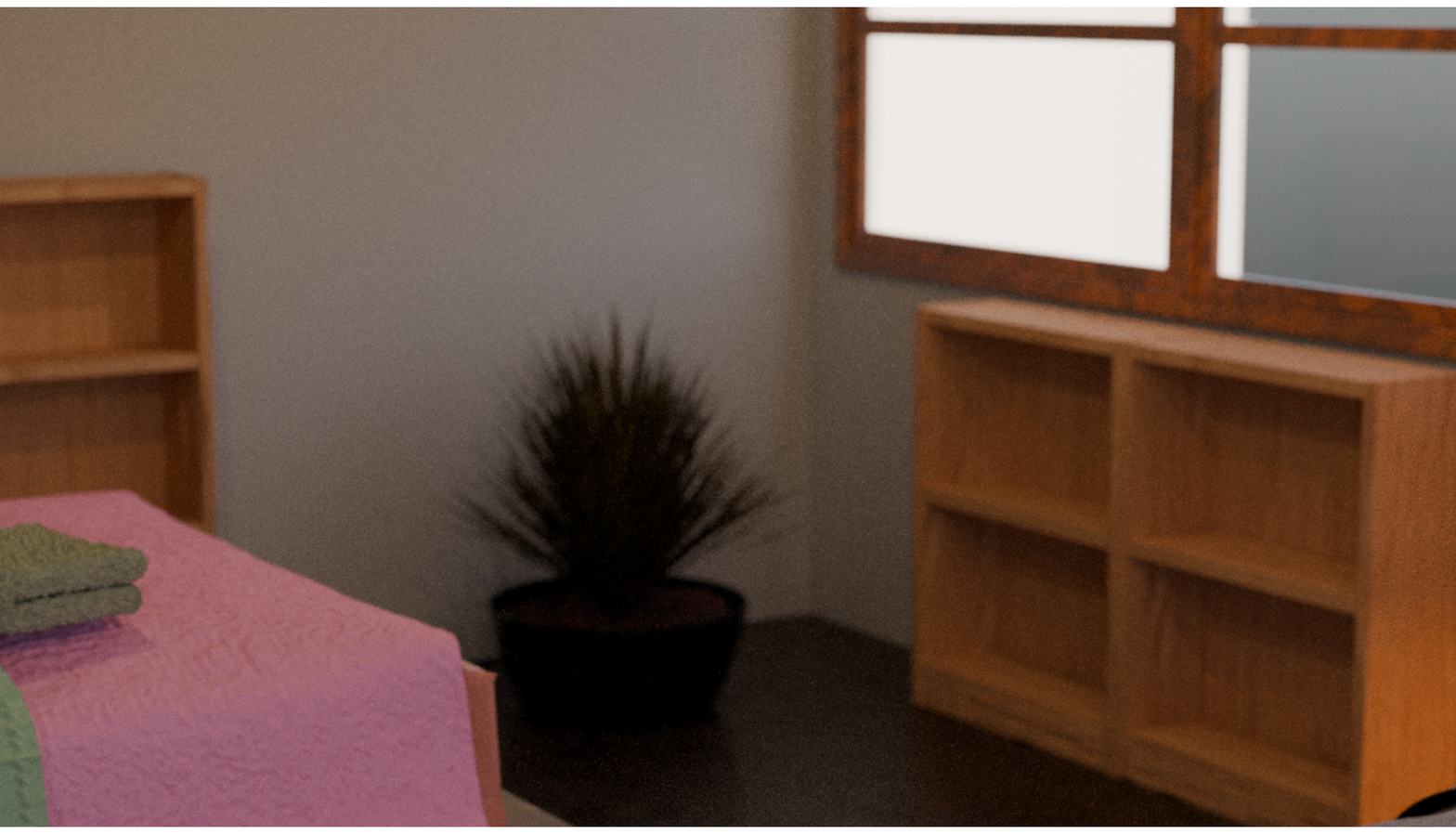}
    \includegraphics[width=0.96\linewidth]{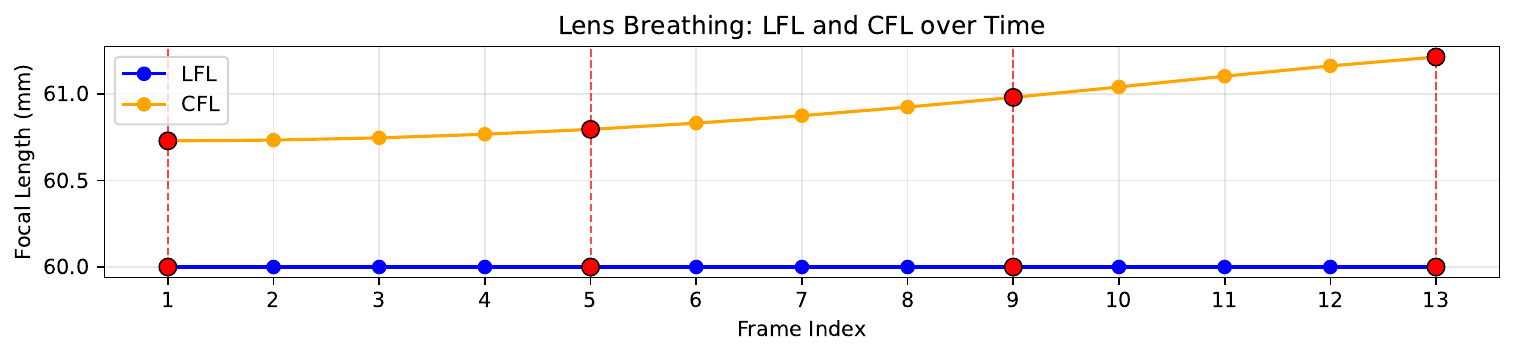}
    \includegraphics[width=0.96\linewidth]{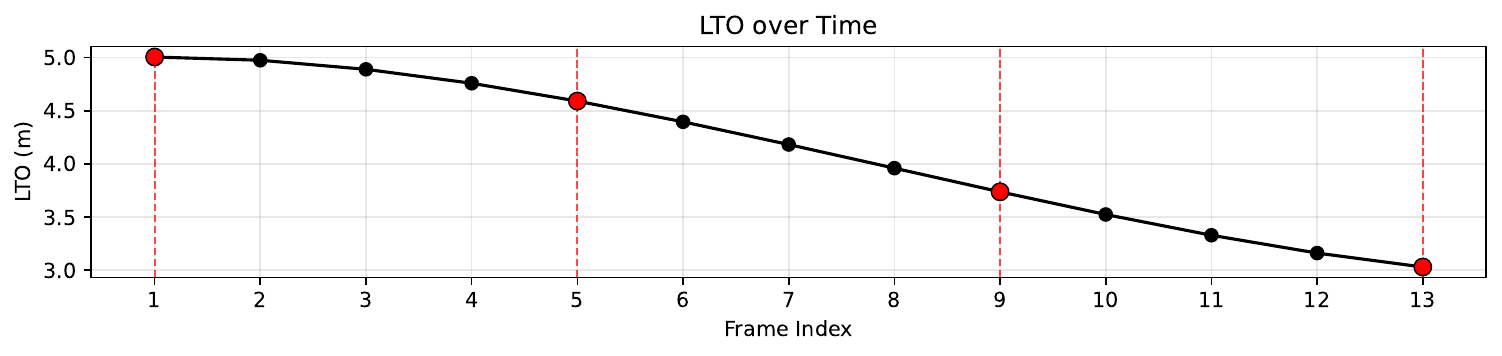}
    \caption{An illustration of lens breathing in synthetic renders. Camera pose and LFL are held constant across all frames while LTO is varied. Because CFL is computed as a function of LFL and LTO, it changes as well, resulting in slight variations in camera FOV. In this example, decreasing LTO causes the camera to zoom in slightly. The effect is most noticeable near the bottom corners of the images.}
    \label{fig:lens_breathing}
\end{figure}

Lens breathing, a phenomenon in real cameras where changes in LTO can cause changes in CFL and therefore camera field of view (FOV), does not occur naturally in Blender's Cycles rendering engine. To add this effect back into our video renders, we first select an LFL value for each video frame, and then compute CFL as a function of LFL and LTO via the thin lens equation. In \cref{fig:lens_breathing}, we provide a toy example illustrating this process and the resulting images. We hold camera pose and LFL constant across all frames while changing the LTO. In response, the CFL changes over time, producing slight variations in the camera FOV. This effect is most noticeable near the image boundaries; we recommend inspecting the bottom corners of the keyframes to observe it clearly.

\section{Additional Details on InFlux++ Synth: Temporal Variation of LFL and LTO}

To produce natural variation in camera intrinsics, we vary LFL and LTO over frames using bounded random walks, as described in the main paper. Here, we provide additional implementation details on the random walk algorithm and the parameters used for InFlux++ Synth.

In our random walk algorithm, the next keyframe value $x_{t+1}$ is sampled repeatedly until it lands within the range $[x_{\min}, x_{\max}]$. In practice, we cap this resampling procedure at 1000 attempts. If no valid sample is obtained after 1000 trials, we keep the previous keyframe value and set $x_{t+1} = x_t$. Empirically, resampling almost never fails if the algorithm step sizes are set properly.

For LTO, instead of performing the random walk directly on the raw LTO value, we perform the random walk over a normalized parameter $\alpha \in [0,1]$ that represents the relative position between the minimum and maximum visible scene depths in the frame. To estimate the visible depth range for each frame, we perform ray tracing with a precomputed scene bounding volume hierarchy (BVH). Specifically, we randomly sample 1000 pixels in the image plane and cast rays from the camera center through these pixels, recording the depth of the first BVH intersection for each ray. To reduce the impact of outliers, we define the minimum and maximum visible scene depths using the 5th and 95th percentile depths of the sampled rays. In the rare case that none of the 1000 sampled pixels returns a valid depth, we instead reuse the minimum and maximum scene depths from the most recent frame with valid estimates.

\begin{figure}[t]
    \centering
    \includegraphics[width=0.96\linewidth]{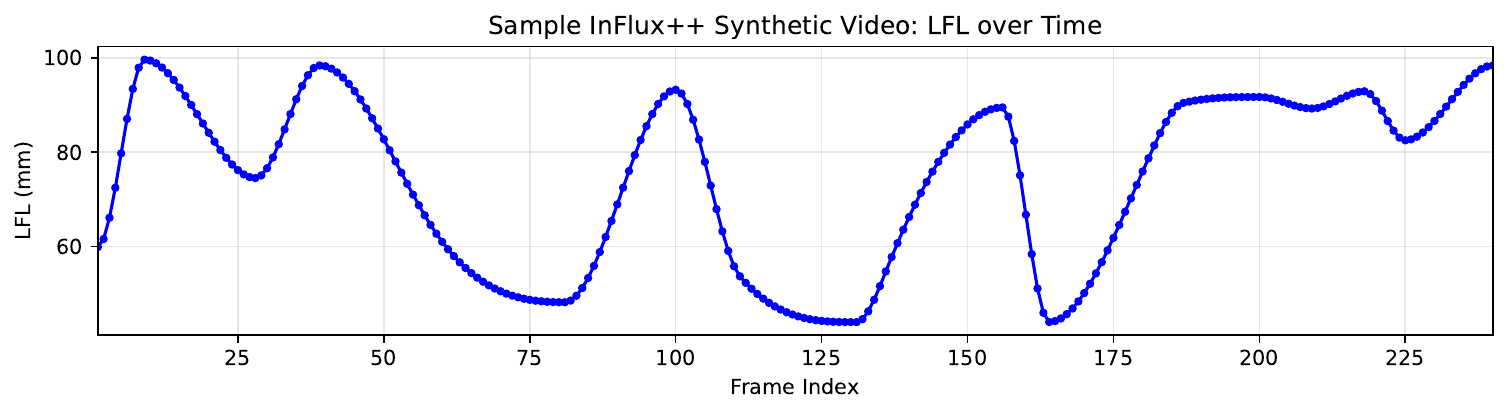}
    \includegraphics[width=0.96\linewidth]{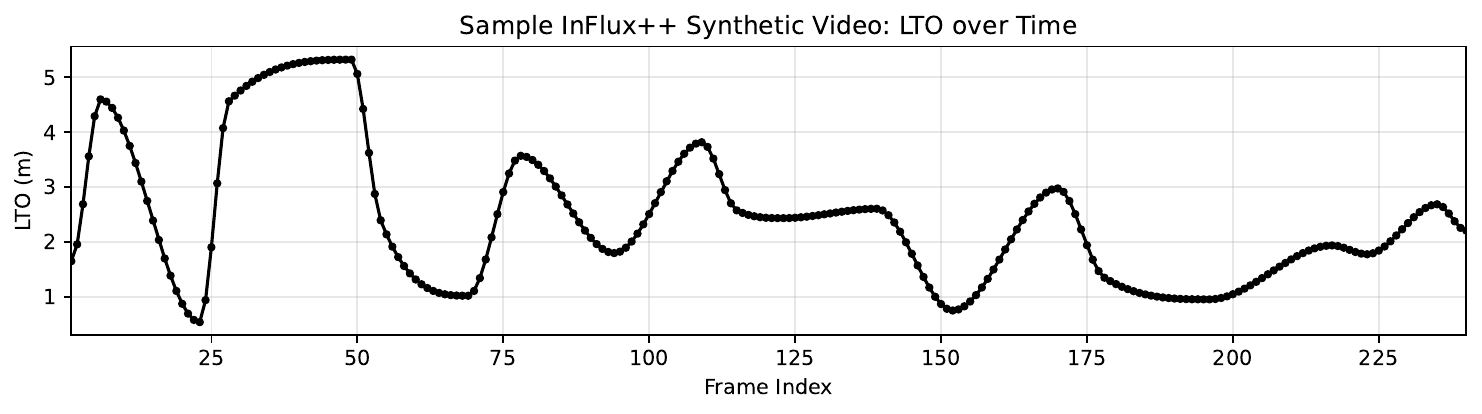}
    \includegraphics[width=0.96\linewidth]{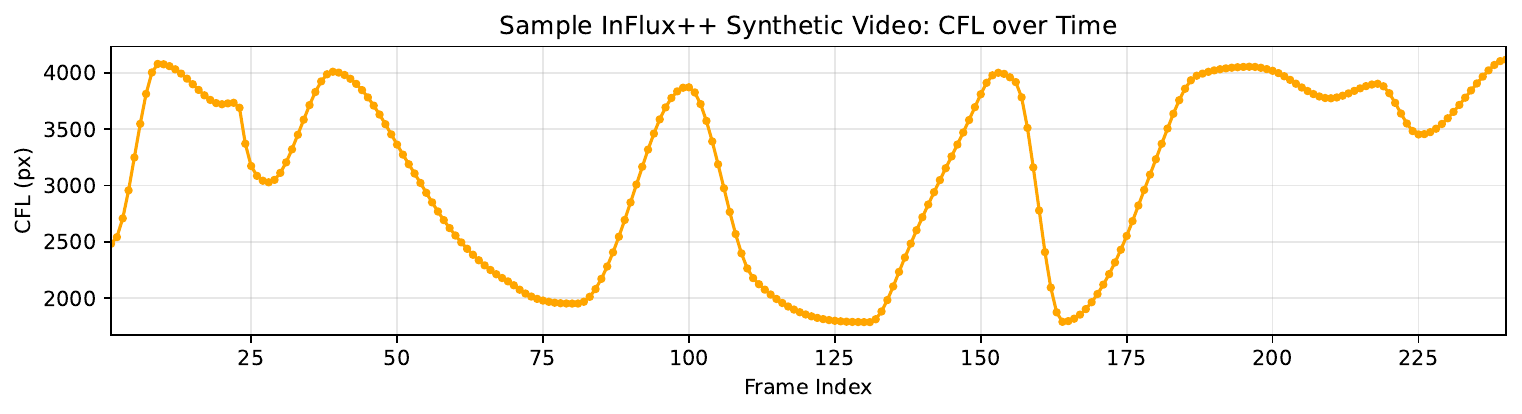}
    \caption{Plots of LFL, LTO, and CFL varying over time for a sample InFlux++ Synth video. LFL and LTO vary naturally over time, so computed CFL also varies naturally.}
    \label{fig:synth_intrinsics}
\end{figure}

By producing natural and smooth changes in LFL and LTO over time, the resulting change in CFL is also natural and smooth. See \cref{fig:synth_intrinsics} for an example of how these parameters change for a sample InFlux++ Synth video.

\section{Additional Details on InFlux++ Synth: Lens Distortion and Augmentations}

\textbf{Lens Distortion.} We add Brown-Conrady~\cite{brown} distortion to InFlux++ Synth training data via our data loader. To do so, we create a sampling method that aims to avoid unrealistic visual artifacts caused by extreme distortion values, and to produce a realistic distribution of distortion effects, including barrel, pincushion, and mustache distortion, conditioned on CFL.

To avoid unrealistic tearing or flipping from distortion, we sample plausible radial displacements at two control points on the image edge---where radial distortion is typically greatest---and solve for the corresponding $k_1$, $k_2$ coefficients. Specifically, we choose the top-left corner $(-W/2, -H/2)$ and the top-center point $(0, -H/2)$. We sample the pixel displacement at the corner $\Delta pxl_c$ from a truncated Gaussian distribution with mean dependent on $\mathrm{CFL}$. We then pick the pixel displacement at the top-center $\Delta pxl_t$ by multiplying $\Delta pxl_c$ by a scale factor from another truncated Gaussian. Here, $\Delta pxl$ is measured in pixels, with positive values indicating outward displacement and negative values indicating inward displacement. Specifically, we use:

\begin{align*}
\Delta pxl_c &\sim \mathcal{N}_{tr} \left( \mu = \mu_{CFL}, \sigma = 50, \min = -50, \max = 80 \right) \\
\Delta pxl_t &\sim \mathcal{N}_{tr} \left( \mu = 0.143, \sigma = 0.15, \min = -0.10, \max = 0.5 \right) \cdot \Delta pxl_c \\
& \text{where} \ \mu_{CFL} = 0.0021 \cdot \text{CFL} + 16.84
\end{align*}

Note that $\mu_{CFL}$ grows linearly with CFL to reflect the greater prevalence of pincushion distortion at high CFL. We hand-tuned the coefficients to produce varied and visually realistic image distortion, though users may adjust them further to suit their training needs.

To achieve these desired pixel displacements, we temporarily ignore tangential distortion and form a system of equations to solve for $k_1, k_2$. Let the corner and top-center control points have normalized radii
\begin{equation*}
r_c = \frac{\sqrt{(H/2)^2 + (W/2)^2}}{\mathrm{CFL}},
\qquad
r_t = \frac{H/2}{\mathrm{CFL}}.
\end{equation*}
Under the Brown-Conrady radial distortion model, we have
\begin{equation*}
\mathbf{x}_d = \mathbf{x}\left(1+k_1r^2+k_2r^4\right),
\end{equation*}
where $\mathbf{x}$ is in normalized image coordinates and $r=\|\mathbf{x}\|$. The resulting radial displacement in normalized coordinates can be expressed as
\begin{equation*}
\Delta r = r\left(k_1r^2+k_2r^4\right) = k_1r^3+k_2r^5.
\end{equation*}
Converting from normalized coordinates to pixels using $\mathrm{CFL}$ gives
\begin{equation*}
\Delta pxl(r) = \mathrm{CFL}\left(k_1r^3+k_2r^5\right).
\end{equation*}
Applying this equation at the corner and top-center control points yields
\begin{equation*}
\begin{bmatrix}
\mathrm{CFL}\, r_c^3 & \mathrm{CFL}\, r_c^5 \\
\mathrm{CFL}\, r_t^3 & \mathrm{CFL}\, r_t^5
\end{bmatrix}
\begin{bmatrix}
k_1 \\
k_2
\end{bmatrix}
=
\begin{bmatrix}
\Delta pxl_c \\
\Delta pxl_t
\end{bmatrix},
\end{equation*}
which can be solved for $k_1$ and $k_2$.

For tangential coefficients $p_1$ and $p_2$, we set both to zero when finetuning AnyCalib~\cite{anycalib}, as the method does not predict tangential distortion. We release a second data loader intended for models that do support tangential distortion. In this configuration, $p_1$ and $p_2$ are sampled from the following distribution:
\begin{align*}
p_1 &\sim \mathcal{N}(0, 0.0001) &
p_2 &\sim \mathcal{N}(0, 0.0001)
\end{align*}
We sample $p_1$ and $p_2$ in this manner because most high-quality cameras have negligible tangential distortion; however, we leave users the option to customize this distribution further.
\\
\\
\textbf{Photometric Augmentations.} After adding distortion, we apply a custom set of photometric augmentations based on AnyCalib's~\cite{anycalib} and GeoCalib's~\cite{geocalib} data loader augmentations. Specifically, we retain their default color and noise perturbations, including random gamma, tone curves, brightness and contrast, color jitter, optional grayscale or sepia conversion, Gaussian noise, JPEG compression, and ISO noise. We use the default parameters and probabilities used for training the AnyCalib $OP_g$ model. However, we disable downscaling, blur, and sharpening operations, because these operations could potentially alter the effective pixel size or interfere with a model's ability to draw signal from blurred pixels computed via ray casting through a thin lens camera model.
\\
\\
\textbf{Geometric Augmentations.} We mostly follow AnyCalib's defaults for image geometric augmentations, but make one modification to aspect ratio randomization. Specifically, AnyCalib samples a target aspect ratio $H/W$ uniformly from $[0.5, 2]$, whose reciprocal endpoints make the range centered on a square aspect ratio. However, training images are not always square, and sampling aspect ratios that are poorly matched to the data distribution can reduce the amount of training signal contributed by pixels near the image corners. These corner regions are especially informative for learning lens distortion, because distortion effects are typically most pronounced far away from the image center. To address this, given an expected training-image aspect ratio $r$, we instead sample from $[0.5, r^2 / 0.5]$. This makes the sampled aspect-ratio range geometrically centered at $r$, while keeping the range within the original bounds of $[0.5, 2]$.

\section{Additional Details on InFlux++ Real: Lenses and Calibration Experiments}

\subsection{Camera Hardware}

Following InFlux~\cite{influx}, we use zoom lenses that record /i Technology lens metadata, enabling access to per-frame LFL and focus distance (FD) values. Our capture setup uses the same types of hardware as InFlux: an ARRI Alexa Mini paired with two zoom lenses, the Canon CINE-SERVO 17–120 mm PL mount (canon17) and the Fujinon Premista 80–250 mm (premista80).

\begin{figure}[t]
    \centering
    \includegraphics[width=0.48\linewidth]{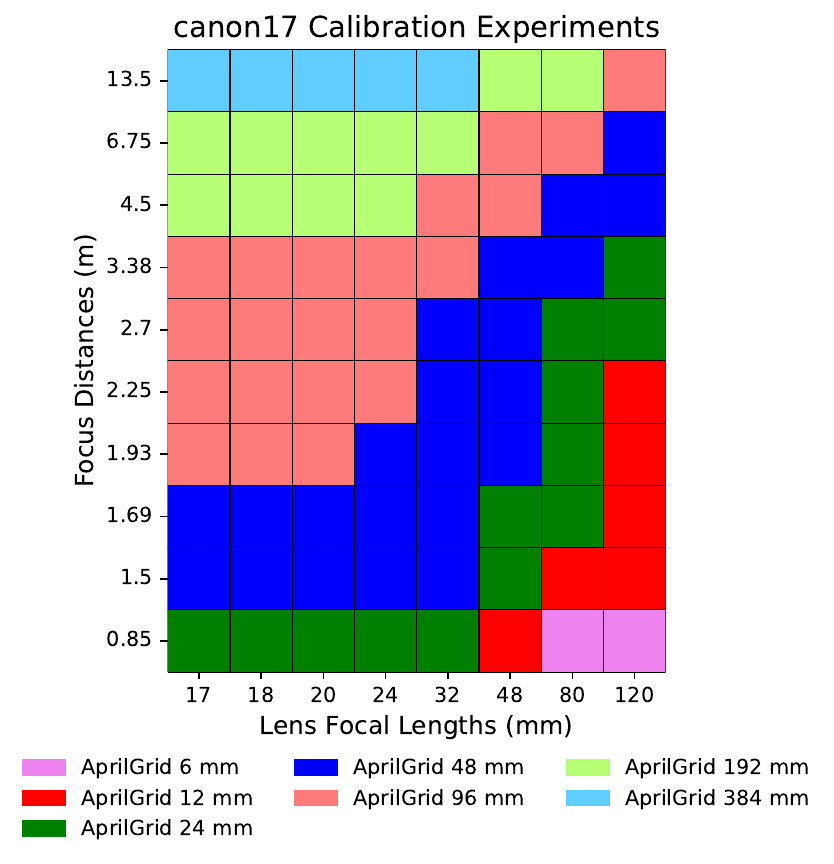}
    \includegraphics[width=0.48\linewidth]{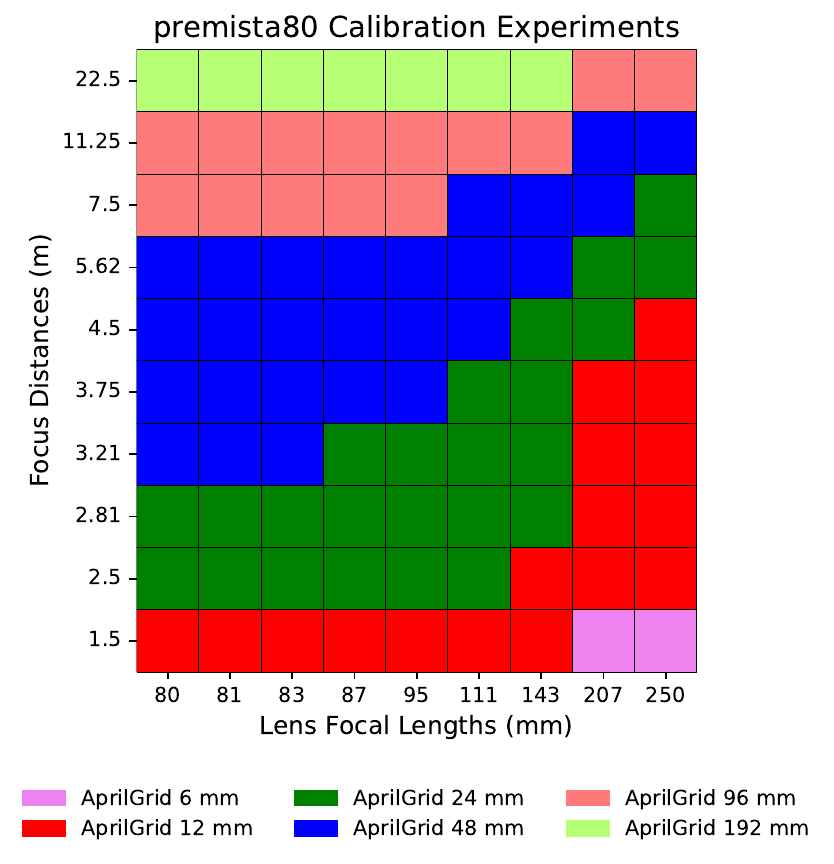}
    \caption{Visualization of the calibration board experiments used to construct the LUTs for canon17 and premista80. Compared to InFlux, we only perform board-based calibration, thereby simplifying the calibration process compared to drone-based calibration.}
    \label{fig:board_settings}
\end{figure}

\subsection{Calibration Experiments}

We use the same lens models as InFlux~\cite{influx}. We calibrate each lens at the same set of LFL and FD values used in InFlux:

\begin{itemize}
    \item canon17 LFLs: 17~mm, 18~mm, 20~mm, 24~mm, 32~mm, 48~mm, 80~mm, 120~mm
    \item canon17 FDs: 0.85~m, 1.69~m, 1.93~m, 2.25~m, 2.7~m, 3.38~m, 4.5~m, 6.75~m, 13.5~m
    \item premista80 LFLs: 80~mm, 81~mm, 83~mm, 87~mm, 95~mm, 111~mm, 143~mm, 207~mm, 250~mm
    \item premista80 FDs: 1.5~m, 2.5~m, 2.81~m, 3.21~m, 3.75~m, 4.5~m, 5.62~m, 7.5~m, 11.25~m, 22.5~m
\end{itemize}

For our calibration experiments, we only perform board-based calibration. Our smaller board patterns are the same as those used in InFlux: an $8 \times 11$ grid of AprilTags~\cite{apriltag} for board sizes of $100 \times 75$~mm, $200 \times 150$~mm, $400 \times 300$~mm, and $800 \times 600$~mm. The corresponding AprilTag square sizes are 6~mm, 12~mm, 24~mm, and 48~mm. Beyond these boards, we introduce three new board patterns: $4 \times 5$ grid of 96~mm tags, $8 \times 11$ grid of 192~mm tags, and $4 \times 6$ grid of 384~mm tags. The 96~mm tag board is constructed by printing the calibration pattern as a sticker and attaching it carefully to a flat $800 \times 600$~mm board. The 192~mm tag and 384~mm tag patterns are projected onto a rigid screen of dimensions $5.45~\text{m} \times 3.06~\text{m}$. For each $(\text{LFL}, \text{FD})$ experiment, we select the largest board that fits within the camera's FOV. See \cref{fig:board_settings} for a visualization of board assignments to calibration experiments.

\begin{figure}[t]
    \centering
    \includegraphics[width=\linewidth]{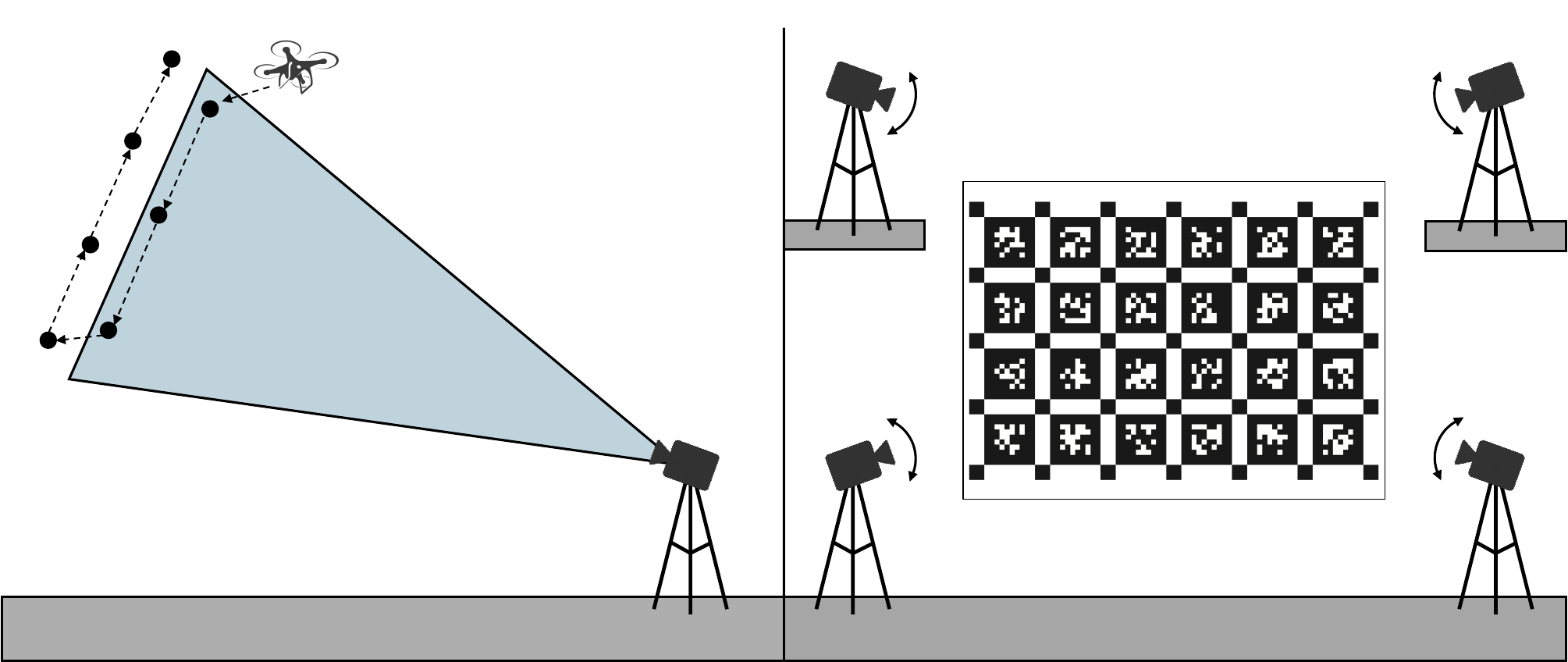}
    \caption{For large FSF experiments, drone-based calibration~\cite{influx} (left) uses a drone with accurate positional tracking as the calibration target and flies it through the static camera's FSF. However, this introduces hardware complexity and potential instability, as environmental disturbances such as wind can cause deviations from intended waypoints. In contrast, our new board-based calibration technique (right) keeps the target fixed and instead moves the camera, sweeping the pattern across the camera FOV to achieve FSF coverage and diverse board orientations. This results in a simpler and more reliable calibration procedure.}
    \label{fig:large_board}
\end{figure}

\begin{figure}[t]
    \centering
    \includegraphics[width=0.45\linewidth]{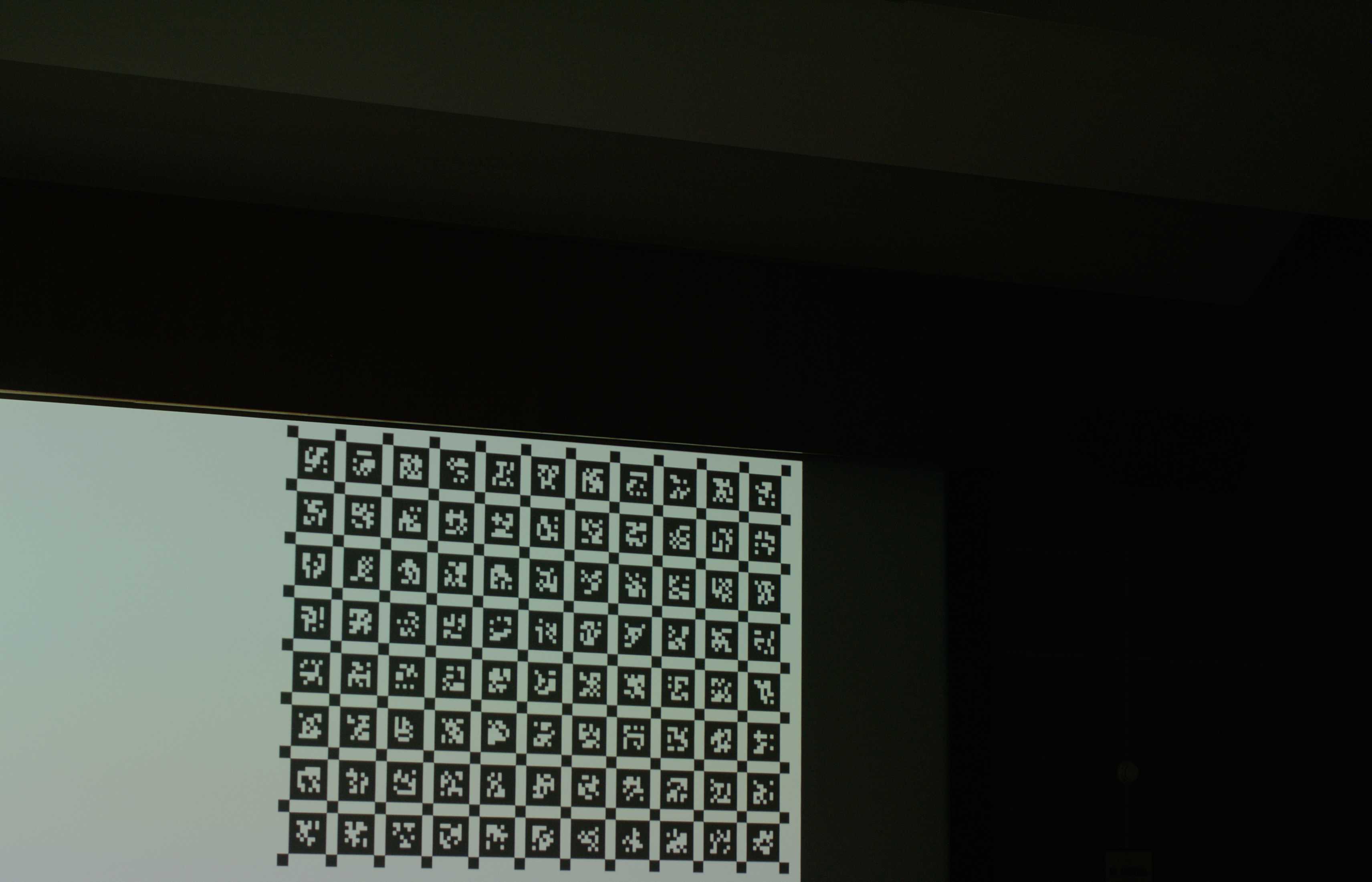}
    \includegraphics[width=0.45\linewidth]{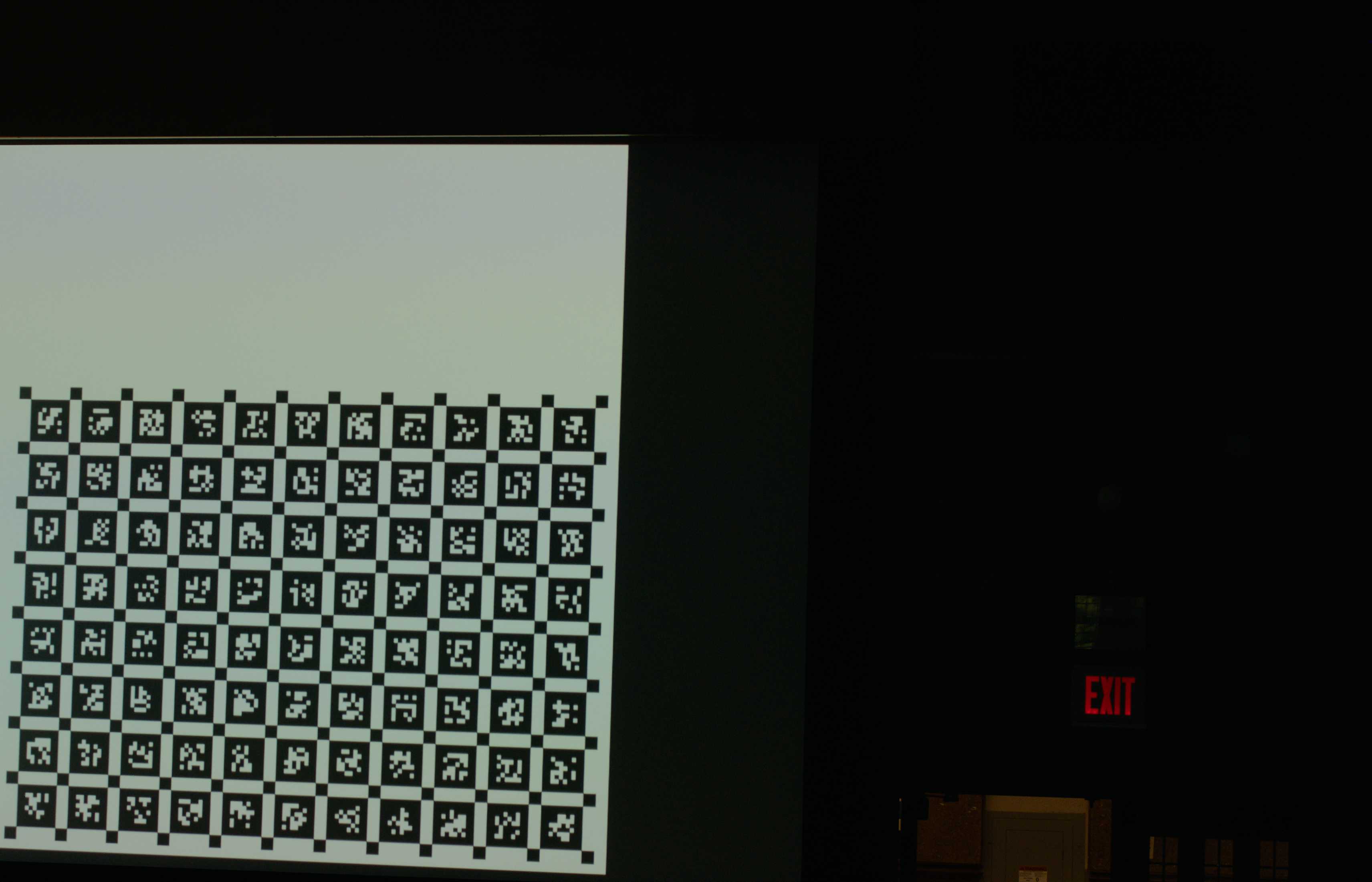}
    \includegraphics[width=0.45\linewidth]{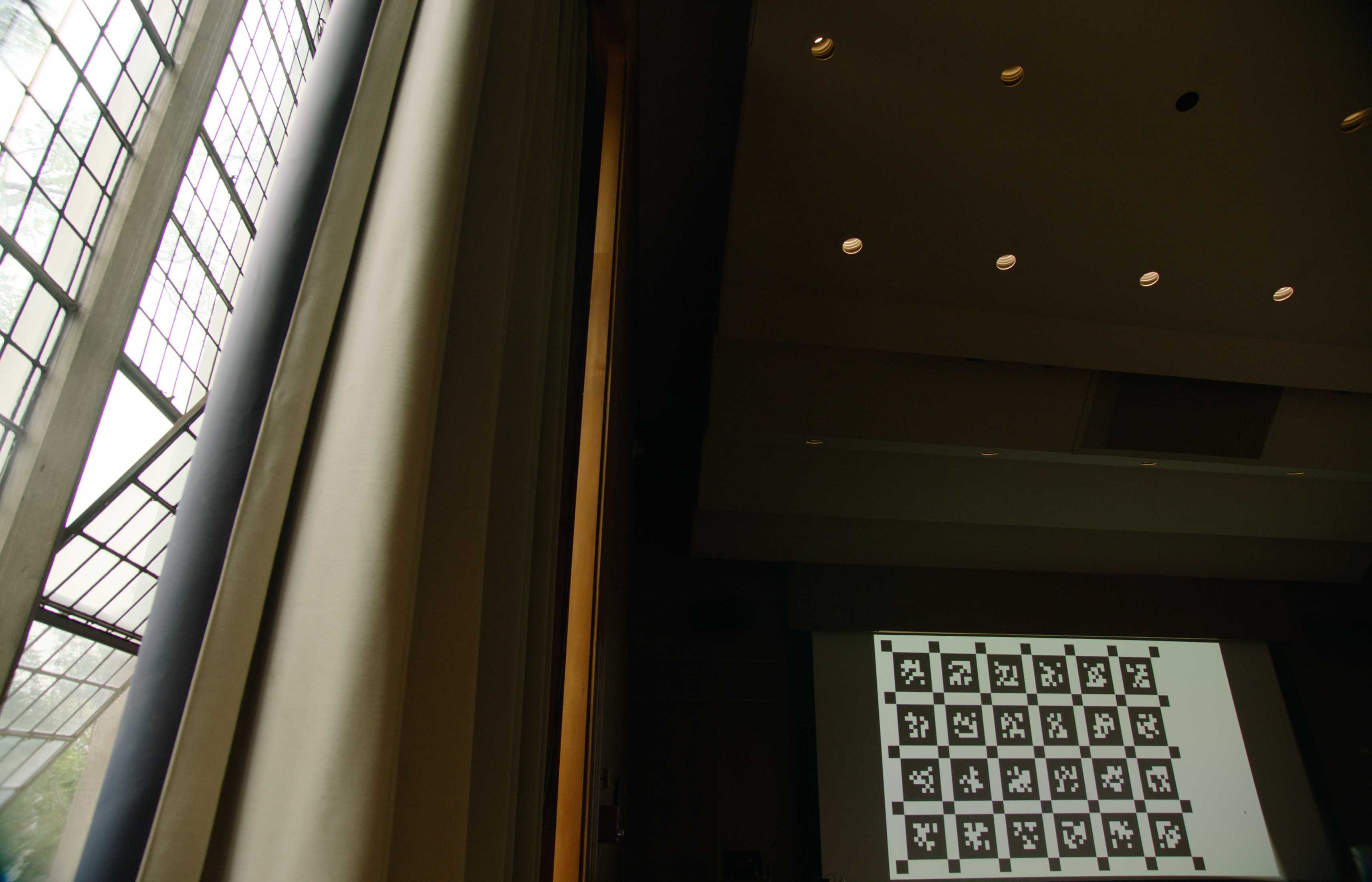}
    \includegraphics[width=0.45\linewidth]{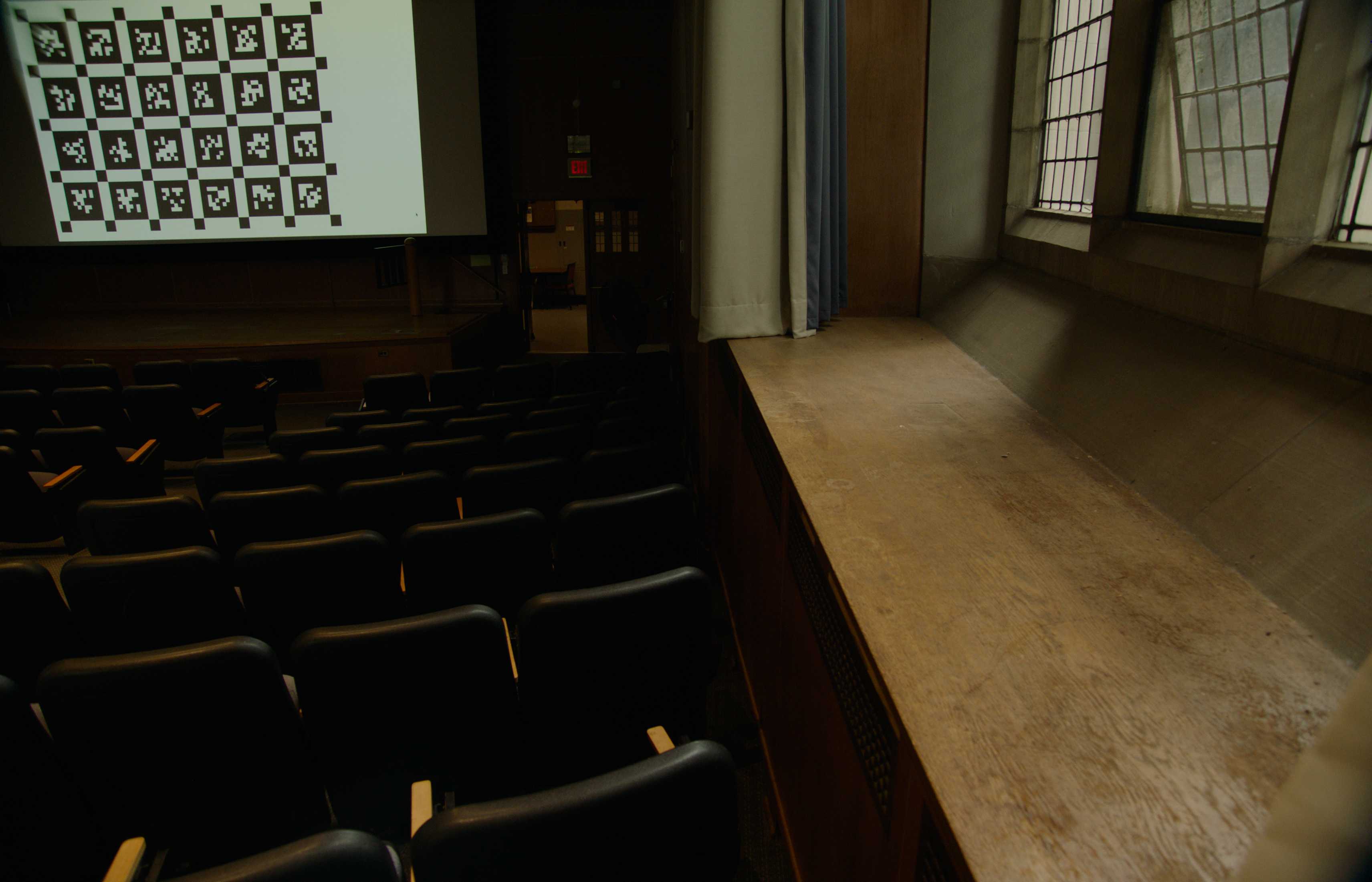}
    \caption{Gallery of calibration images taken of large static board targets in a multi-story lecture hall. The images shown are taken at four positions in the room: left and right sides of the room on the ground floor, and left and right sides of the room on the balcony. We also pan and tilt the camera to capture the calibration board at multiple angles to excite axes of rotation.}
    \label{fig:calibration_gallery}
\end{figure}

To perform calibration with a large static board target, we project our calibration pattern onto a rigid screen located in a multi-story lecture hall with a balcony level. We capture calibration images from four positions: the left and right sides of the room on the ground floor, and the left and right sides of the balcony. At each position, we pan and tilt the camera so that the calibration target moves throughout the entire camera FOV. Filming from multiple corners of the room and rotating the camera at each position produces a wide range of board viewing angles, thereby exciting axes of rotation required for downstream camera calibration. Although the balcony does not extend across the entire length of the room, at higher FDs the camera's depth of field is sufficiently large that the board remains in focus even when it is not positioned exactly at the desired FD. For FDs significantly shorter than the extent of the balcony, we instead film at additional 2-3 positions on the ground floor. Compared to drone-based calibration~\cite{influx} for large FOV spatial footprint (FSF) experiments, our new board-based calibration technique is significantly simpler and more reliable. See \cref{fig:large_board} for a comparison between the drone-based and board-based techniques. See \cref{fig:calibration_gallery} for a gallery of calibration images taken at corners of the multi-story lecture hall.

\section{Additional Details on InFlux++ Real: Face and License Plate Blurring}

To automate the process of blurring faces and license plates in InFlux++ Real, we utilize the RetinaFace~\cite{retinaface} and EgoBlur~\cite{egoblur} codebases, respectively. However, the size of faces and license plates can vary greatly due to changes in zoom within our real-world benchmark videos, making it difficult to consistently detect all instances using a single detection scale.

To address this, we build multi-scale detection systems around both RetinaFace and EgoBlur. For face detection, we run RetinaFace at three input resolutions: $640 \times 640$, $1280 \times 1280$, and $2560 \times 2560$. Detections from all scales are pooled together and filtered using non-maximum suppression (NMS) with an intersection over union (IoU) threshold of 0.5 to remove duplicate overlapping detections.

For license plate detection, we apply EgoBlur to both the full image and to a $3 \times 3$ grid of overlapping image tiles. Each tile covers a region of size $\frac{H}{2} \times \frac{W}{2}$ and is upsampled by a factor of two. After applying EgoBlur to each image, we retain detections with confidence greater than 0.985. Afterwards, we apply NMS with an IoU threshold of 0.5 to remove duplicates before blurring the resulting license plate regions.

\section{Additional Details on InFlux++ Real: Degree of Parallax Comparison}

We provide a qualitative comparison of parallax in InFlux++ Real and InFlux. For each video, we assign one of three labels based on the apparent amount of camera translation. We label a video as low parallax if there is no visually discernible evidence of camera translation. We label it as medium parallax if camera translation is visible but limited in magnitude. We label it as high parallax if the camera undergoes substantial translation over the course of the video. Compared to InFlux, InFlux++ Real contains substantially more high-parallax videos. See \cref{tab:parallax_comparison} for the resulting category counts.

\section{Additional Details on InFlux++ Real: Lens and Intrinsics Diversity}

\begin{figure}[t]
    \centering
    \includegraphics[width=0.48\linewidth]{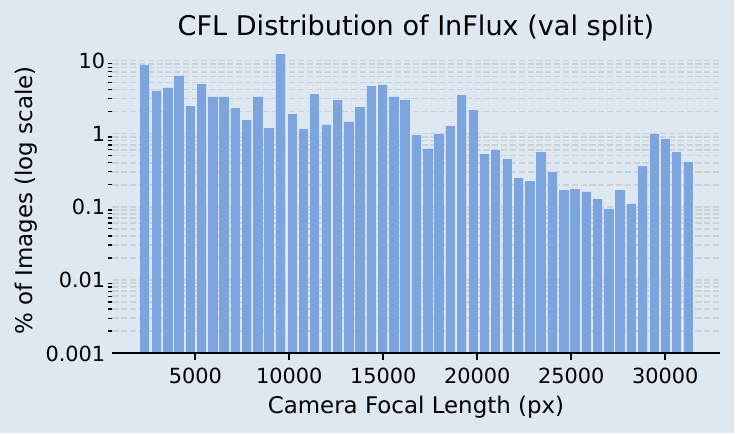}
    \includegraphics[width=0.48\linewidth]{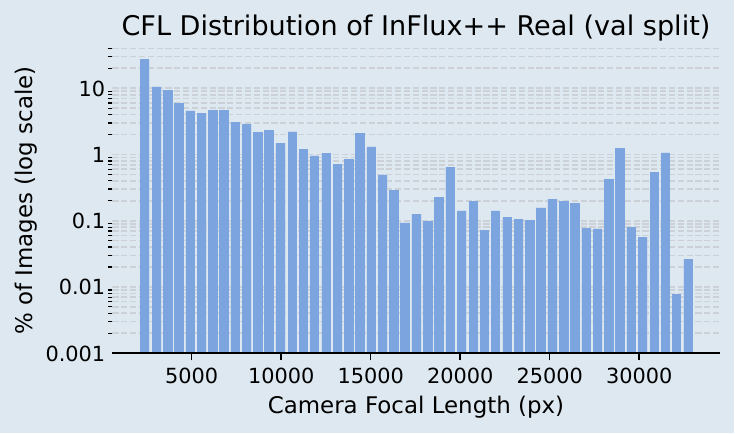}
    \includegraphics[width=0.48\linewidth]{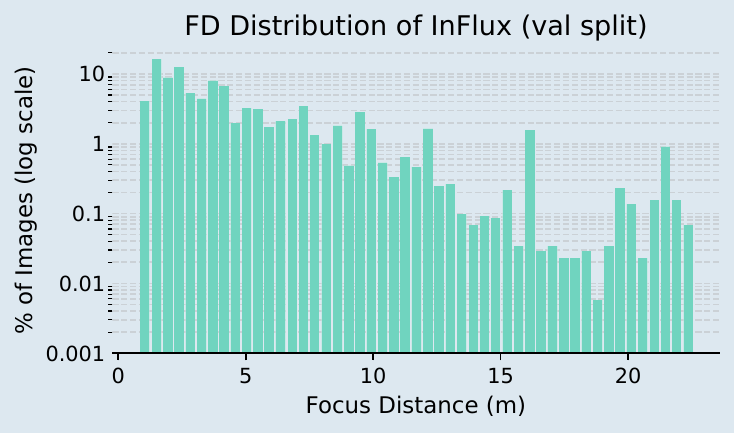}
    \includegraphics[width=0.48\linewidth]{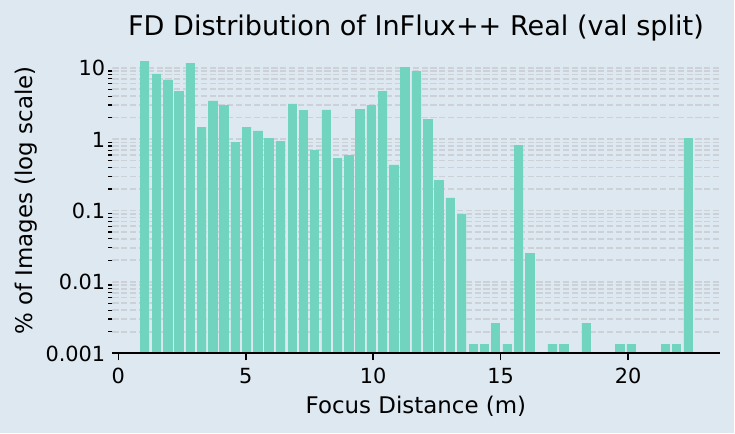}
    \caption{Histograms of per-frame CFL and FD in the validation splits of InFlux and InFlux++ Real. Here, CFL is computed as the average of $f_x$ and $f_y$.}
    \label{fig:real_hist_plots_cfl_fd}
\end{figure}

For InFlux++ Real, we use the same physical premista80 lens as~\cite{influx}. Our canon17 lens is also the same physical lens used in~\cite{influx}, but it underwent lens repairs since its use in~\cite{influx}, resulting in a different lookup table (LUT). The repaired canon17 exhibits a larger principal point offset, broadening the range of off-center $c_x$ and $c_y$ values in InFlux++ Real relative to InFlux. See \cref{fig:real_hist_plots_cfl_fd} for a side-by-side comparison of CFL and FD distributions of the validation splits of InFlux and InFlux++ Real.

\begin{figure}[t]
    \centering
    \includegraphics[width=0.48\linewidth]{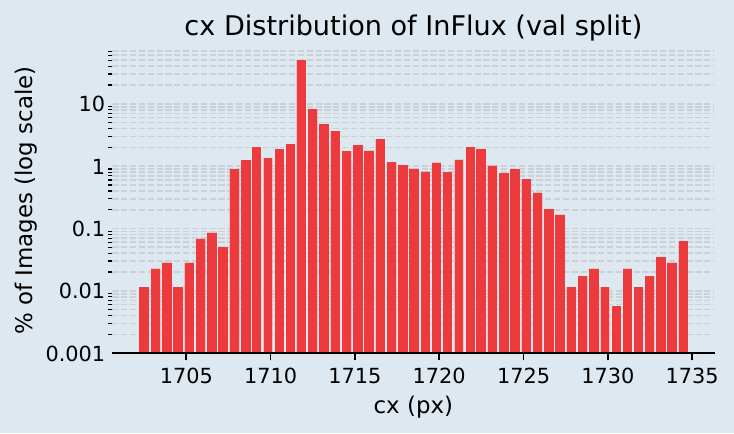}
    \includegraphics[width=0.48\linewidth]{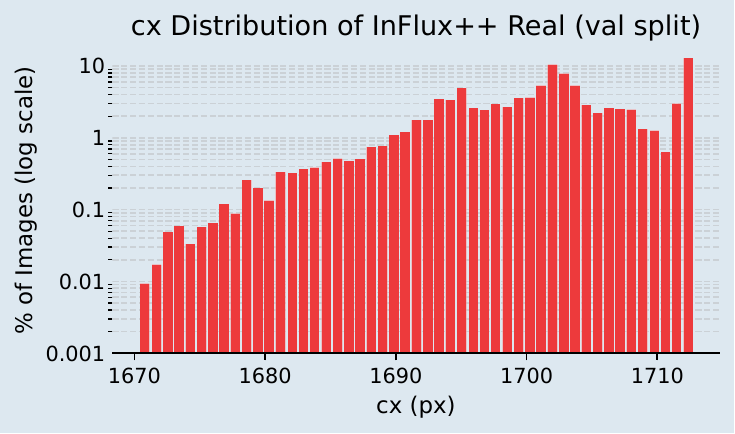}
    \includegraphics[width=0.48\linewidth]{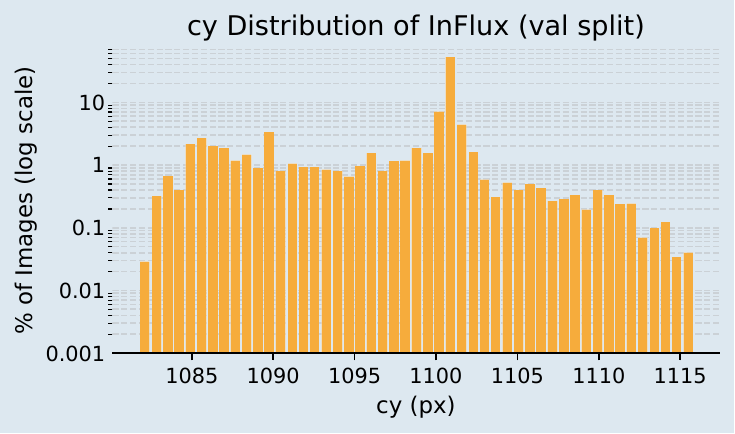}
    \includegraphics[width=0.48\linewidth]{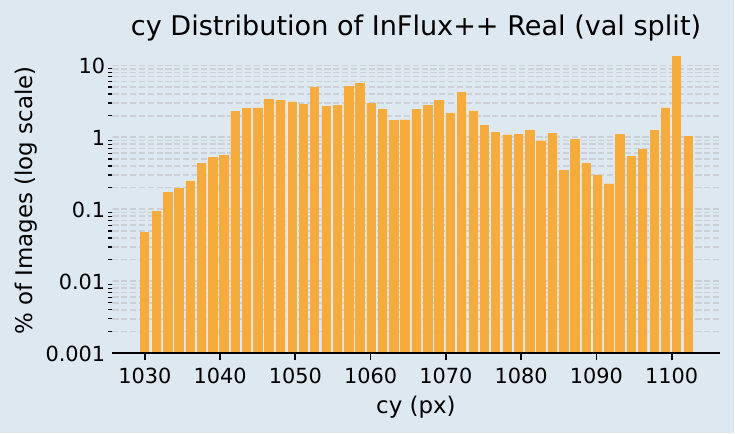}
    \caption{Histograms of per-frame $c_x$ and $c_y$ in the validation splits of InFlux and InFlux++ Real.}
    \label{fig:real_hist_plots_cxcy}
\end{figure}

In many benchmark videos filmed for InFlux++ Real, we closely track nearby subjects as they move and keep them in sharp focus. In contrast, the focus ring is varied in a more arbitrary fashion for~\cite{influx}. As a result, InFlux++ Real exhibits shorter FD values than those found in InFlux. See \cref{fig:real_hist_plots_cxcy} for a side-by-side comparison of $c_x$ and $c_y$ distributions of the validation splits of InFlux and InFlux++ Real.

\begin{table}[t]
  \centering
  \small
  \caption{Qualitative comparison of parallax in InFlux and InFlux++ Real. InFlux++ Real contains substantially more high-parallax sequences.}
  \label{tab:parallax_comparison}
  \setlength{\tabcolsep}{6pt}
  \begin{tabular}{l rr}
    \toprule
    \textbf{Category} & \textbf{InFlux} & \textbf{InFlux++ Real} \\
    \midrule
    \textbf{Low Parallax} & 324 & 119 \\
    \textbf{Medium Parallax} & 35 & 28 \\
    \textbf{High Parallax} & 27 & 183 \\
    \bottomrule
  \end{tabular}
\end{table}

\section{Additional Details on Evaluation Metrics and Procedure}

\subsection{Corrected EPE 3D Point Visibility Filter}

\begin{figure}[t]
    \centering
    \includegraphics[width=\linewidth]{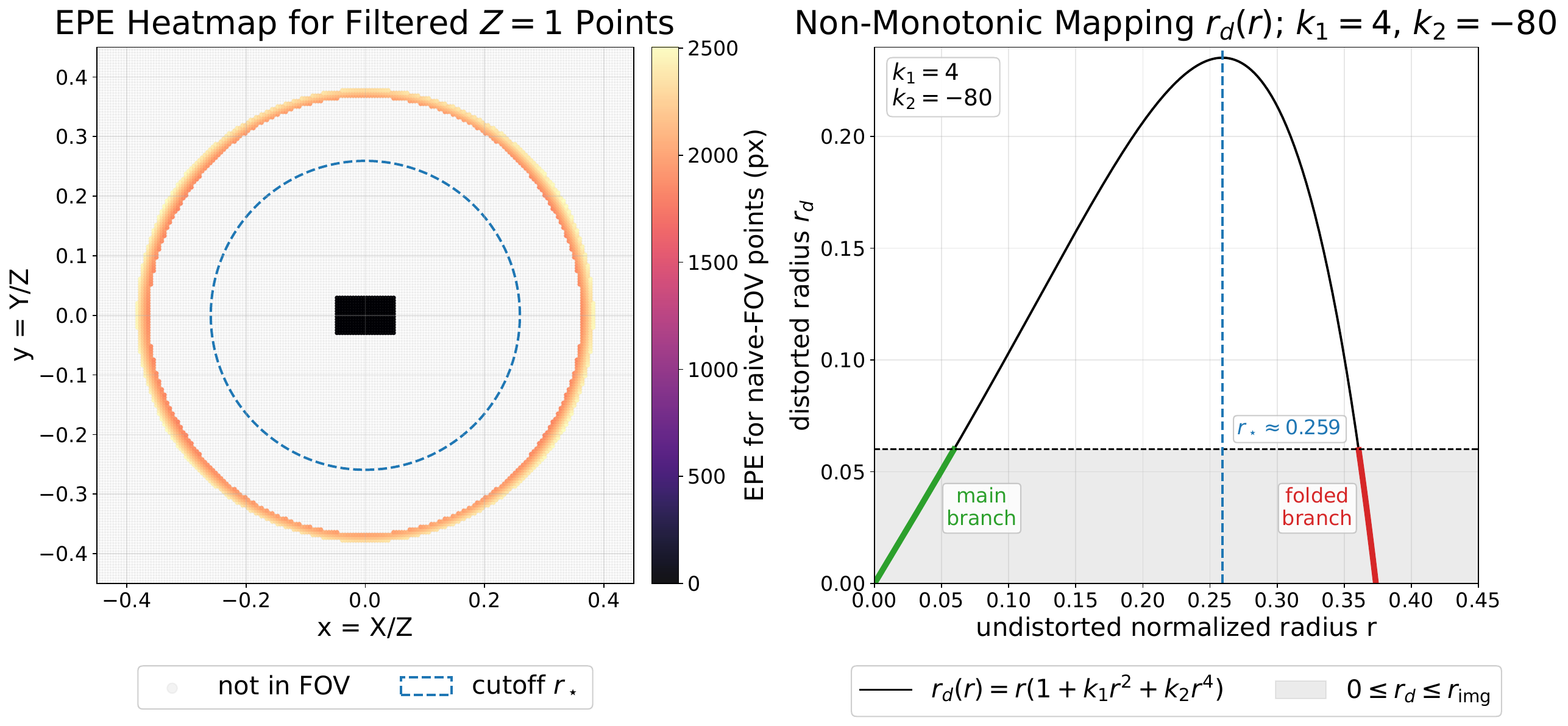}
    \caption{Inherited visibility filter applied to $Z = 1$ points using ground truth intrinsics: $f_x=f_y=30000$, $(k_1,k_2)=(4,-80)$. \textit{Left}: the inherited filter incorrectly keeps a far-off-axis ring, which has large EPE under $\Delta k_2=-10$. \textit{Right}: the ring lies beyond the cutoff $r_*$ where $r_d(r)$ folds back. The corrected filter only keeps points within image bounds and the physically valid support, $r<r_*$.}
    \label{fig:epe_filter}
\end{figure}

The endpoint error (EPE) computation protocol introduced by~\cite{influx} can incorrectly retain far-off-axis 3D points that are folded back into the image by Brown-Conrady~\cite{brown} distortion, even though they are not physically visible to a real lens; we correct this by restricting EPE visibility to the main monotonic support of the radial mapping. From a set of 100K 3D points sampled from~\cite{ETH-3D}, EPE selects which to evaluate based on visibility to the ground truth camera. The original filter from~\cite{influx} projects each point with ground truth intrinsics, applies distortion, and treats points landing inside image bounds as visible, but \cref{fig:epe_filter} shows a toy example where this approach can incorrectly accept far-off-axis points that a real lens would not see. To detect such points, we note that for a real lens with radial distortion, distorted radius $r_d(r)$ should be monotonically increasing over the visible support, where $r^2=(X/Z)^2+(Y/Z)^2$ and $r_d(r)=r(1+k_1r^2+k_2r^4)$. Beyond this support, the model can behave non-physically, folding far-off-axis points back in-bounds as polynomial artifacts. We correct the filter by setting cutoff radius $r_*$ to the first positive root of $r_d'(r)=1+3k_1r^2+5k_2r^4$, if one exists, and keeping only in-bounds points with $r<r_*$. This removes far-off-axis points that fold back into the image while preserving physically valid visible points.

\subsection{LUT-Reliable EPE Recall}

To ensure EPE recall on InFlux++ Real and InFlux is computed only on frames with reliable ground truth intrinsics from LUT interpolation, we formulate LUT-reliable EPE recall, a filtered metric that uses leave-one-out (LOO) validation over each LUT to identify regions reliable for evaluation at each EPE pixel threshold $T$. In each LOO trial, we hold out one LUT calibration data point, estimate its intrinsics by interpolating among its remaining neighbors, and compare the interpolated intrinsics to the measured intrinsics at the held-out sample using EPE recall@$T$~pixels (px). For each threshold $T$, we mark a LUT vertex as reliable if its LOO EPE recall@$T$~px is greater than or equal to $\tau=0.95$. When computing LUT-reliable EPE recall@$T$~px, we include only frames whose intrinsics come from a LUT region that has all vertices reliable at threshold $T$. We use LUT-reliable EPE recall for all reported EPE metrics.

\begin{figure}[t]
    \centering
    \includegraphics[width=0.32\linewidth]{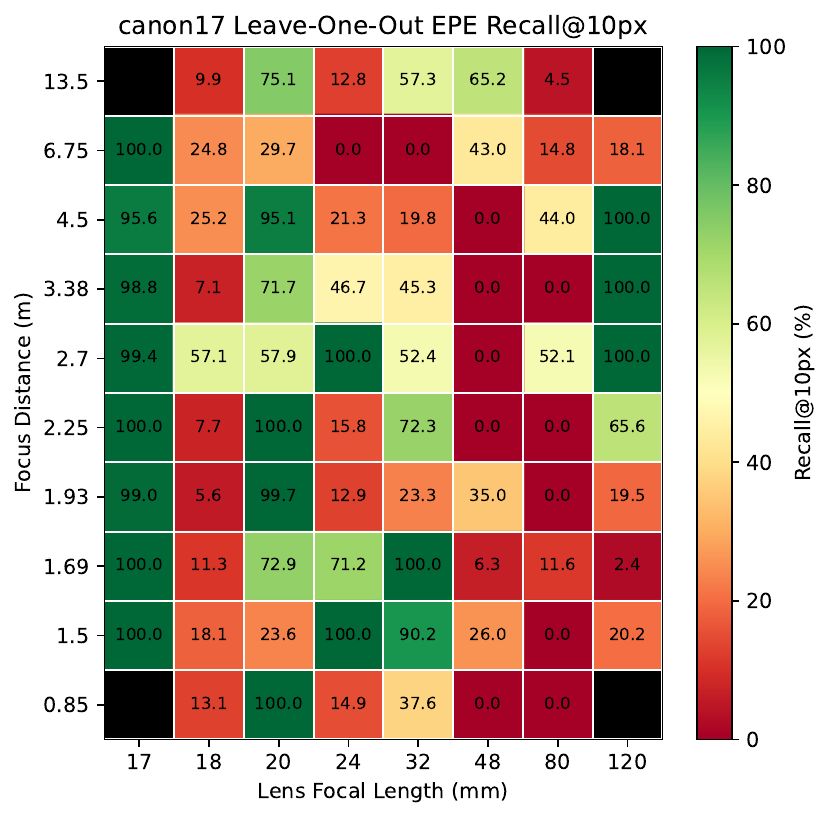}
    \includegraphics[width=0.32\linewidth]{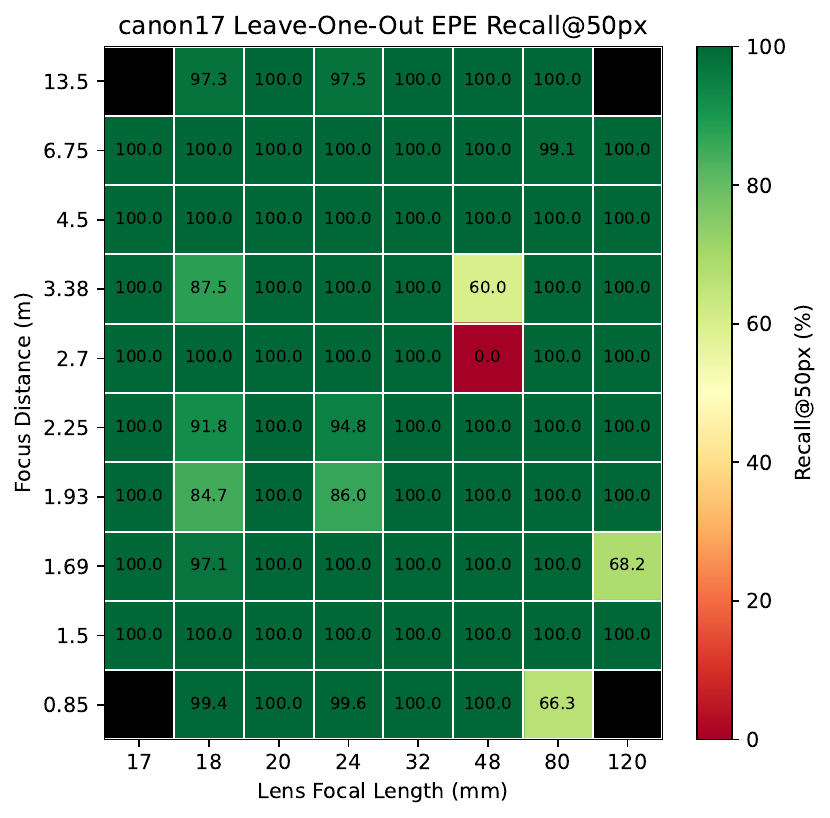}
    \includegraphics[width=0.32\linewidth]{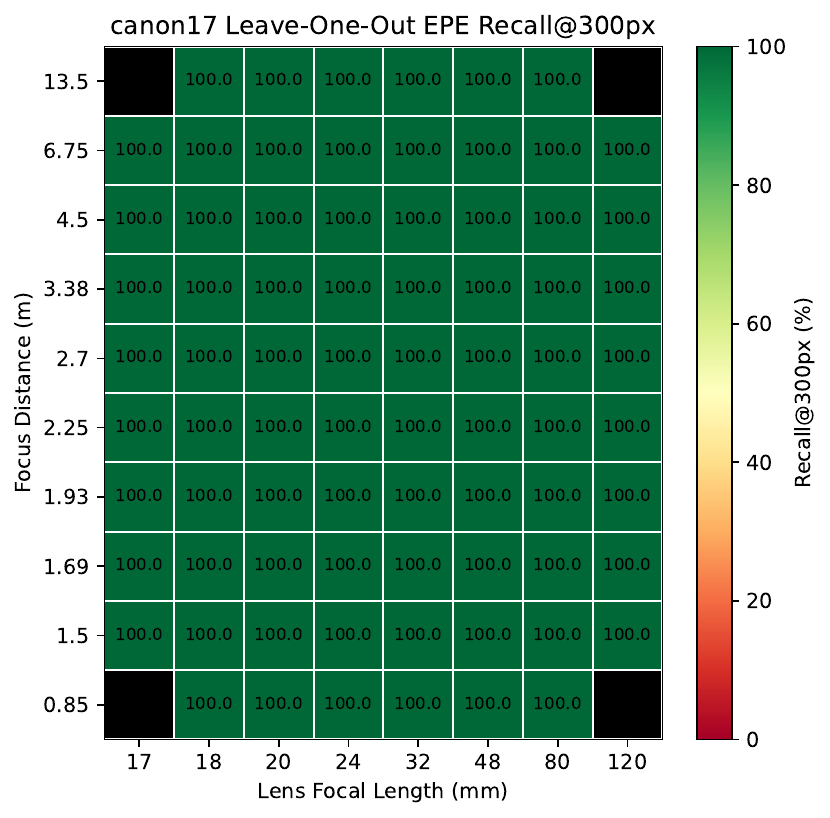}
    \includegraphics[width=0.32\linewidth]{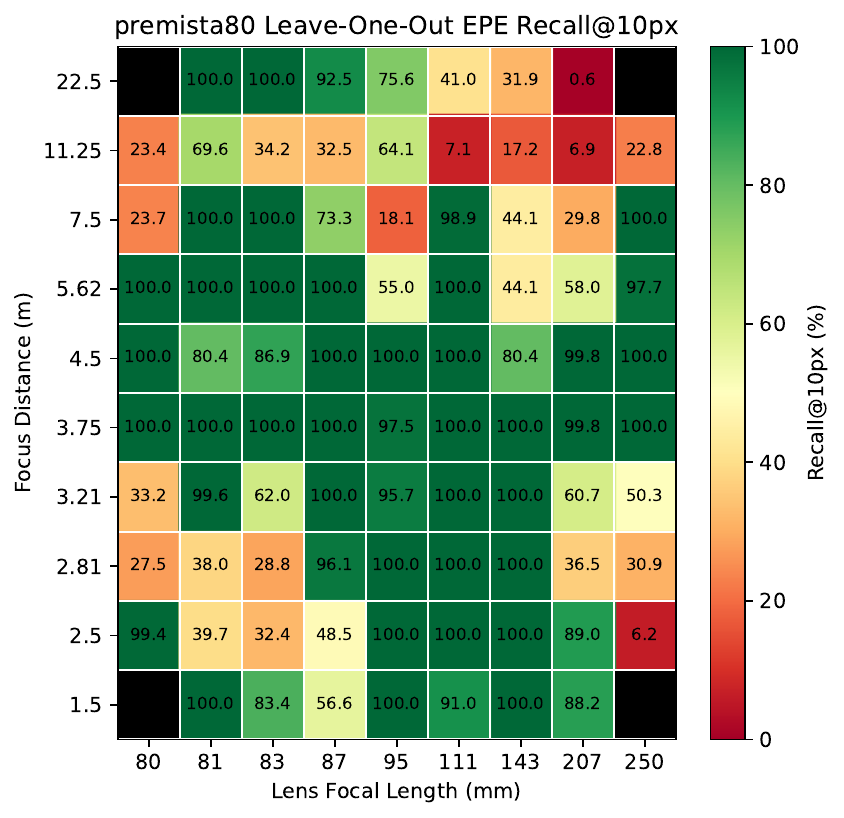}
    \includegraphics[width=0.32\linewidth]{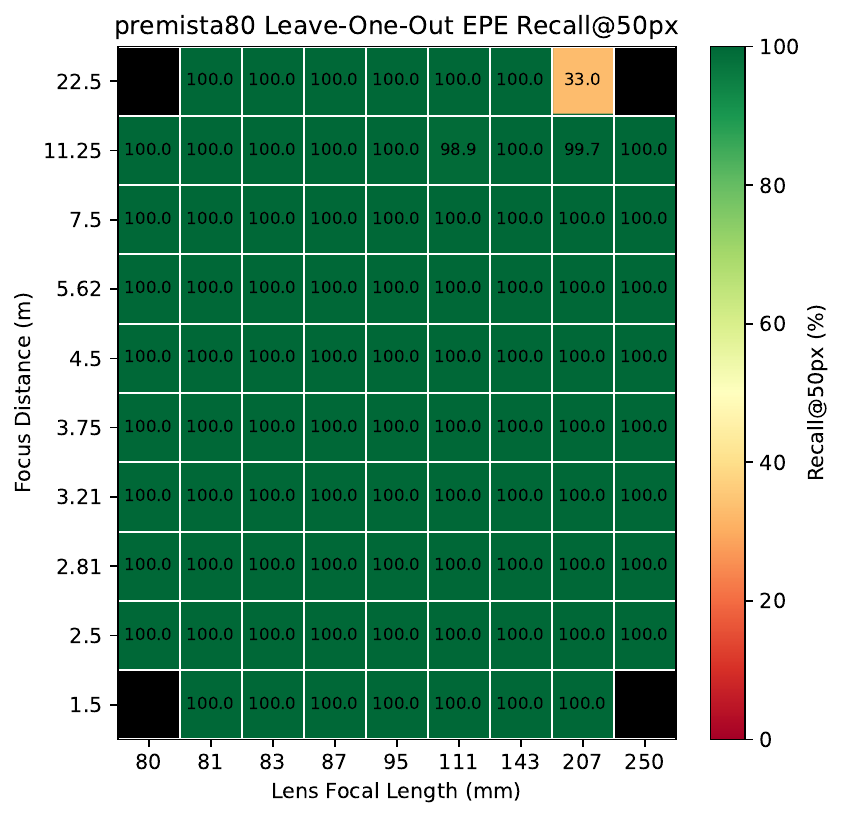}
    \includegraphics[width=0.32\linewidth]{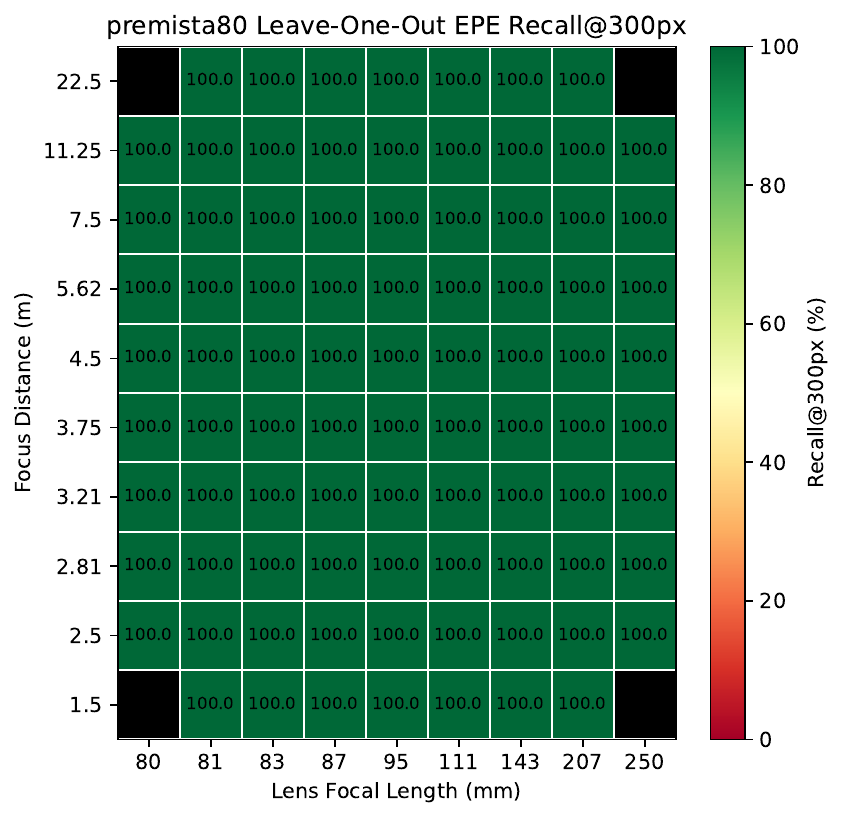}
    \caption{LOO EPE recall heatmaps for canon17 and premista80 lenses at 10/50/300~px.}
    \label{fig:loo_epe_heatmaps}
\end{figure}

Our benchmark currently supports LUT-reliable EPE recall for all integer thresholds $T$ from 1 to 300 pixels, inclusive. \cref{fig:loo_epe_heatmaps} shows how the reliability of LUT vertices changes as we vary $T$ for both canon17 and premista80 lenses. LOO EPE recall@10~px identifies some threshold-specific unreliable LUT vertices. For LOO EPE recall@50~px and recall@300~px, most LUT vertices are reliable.

\section{Additional Details on AnyCalib Finetuning}

\subsection{Training Data Details}

For training data, we use the portion of InFlux++ Synth data without depth and surface normal annotations, which consists of 1023 videos and 245K+ frames in total. We further split this into training and validation sets with an 80/20 ratio, and we keep all frames of a video together in the same split. The training split contains 826 videos for a total of 198,240 images; the validation split contains 197 videos for a total of 47,280 images.

\subsection{Training Details and Hyperparameters}

We initialize AnyCalib from the pretrained checkpoint \texttt{anycalib\_gen.pt} available on Hugging Face, and maintain the original AnyCalib architecture. Because our training images have a resolution of $1280 \times 720$, we randomize aspect ratio by sampling $H/W$ from the range $[0.5, 0.633]$.  We finetune with a global batch size of 48 and optimize the standard AnyCalib l1-z1 ray loss using AdamW. We use a base learning rate of \(7.5 \times 10^{-6}\). Following AnyCalib's default learning rate scaling, the DINOv2 backbone is trained at \(0.1\times\) the base rate, and the decoder and calibration head are trained at the full base rate. The learning rate is stepped once per iteration using a SequentialLR schedule. We use a linear warmup from \(10^{-3}\) times the base learning rate over the first half epoch, followed by MultiStepLR decay with factor \(\gamma = 0.3\) at epochs 4 and 8. Training was performed on 8 NVIDIA RTX 3090 GPUs, and 15 epochs of training took approximately 6 days to complete.

\begin{figure}[t]
    \centering
    \includegraphics[width=\linewidth]{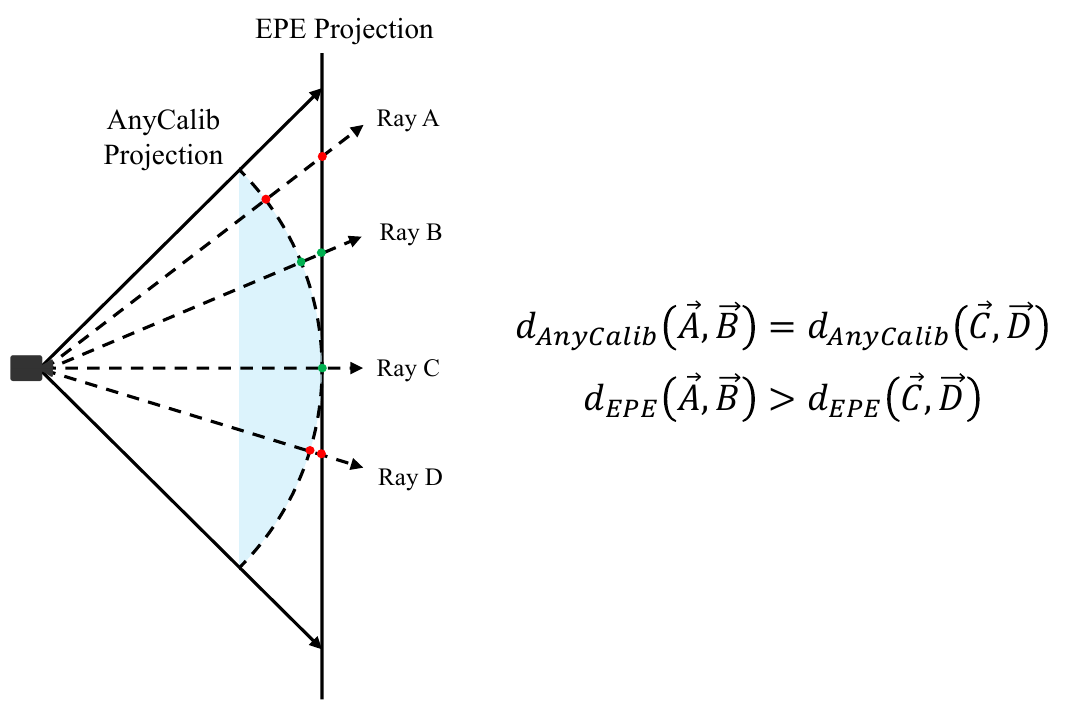}
    \caption{Illustration of the projection mismatch between the AnyCalib loss and the EPE metric. The AnyCalib loss corresponds to the angular distance between rays on the unit sphere, while EPE measures Euclidean distance after projection onto the $z = 1$ plane. Therefore, identical spherical AnyCalib distances can correspond to different EPE distances, with rays near image boundaries incurring larger EPE.}
    \label{fig:finetuning_loss}
\end{figure}

\subsection{AnyCalib Training Objective Analysis}

After finetuning AnyCalib with InFlux++ Synth data, we observe that recall for CFL percent error improves, while recall for EPE slightly decreases. One possible explanation is that the training objective used by AnyCalib may place less emphasis on accurately predicting ray directions near the image boundaries. Since distortion effects are most pronounced near image edges, inaccuracies in these regions may have a negative impact on EPE despite improvements in pinhole camera-related intrinsics.

To predict camera intrinsics, AnyCalib uses a neural network that learns per-pixel ray directions; afterwards, it optimizes a set of intrinsics parameters that best fit the ray directions. To optimize its neural network, it defines a custom ray direction loss. Given two rays $A$ and $B$, AnyCalib first projects them onto the unit sphere, parameterizes their deviation relative to the optical axis intersection point $(0,0,1)$ along the surface of the sphere, and applies an $\ell_1$ loss. Intuitively, this corresponds to measuring distance along the surface of the unit sphere. In contrast, the EPE metric projects rays onto the $z=1$ plane and computes the $\ell_2$ distance between the resulting 2D points. See \cref{fig:finetuning_loss} for a visualization.

Because these two projections measure ray differences differently, equal angular distances on the unit sphere do not necessarily correspond to equal distances under the EPE metric. In particular, pairs of rays near the image boundaries produce larger EPE differences than rays near the image center when holding angular deviation constant. Consequently, optimizing the AnyCalib loss implicitly places greater emphasis on accurate ray directions near the image center, while allowing larger errors near the image edges. This may explain why distortion parameters estimated by AnyCalib are less accurate, leading to degraded EPE performance despite improvements in CFL prediction.

\section{Data Release and Licensing Details}

We publicly release data from both components of InFlux++, available for download on our project website. We release a data loader for InFlux++ Synth that adds lens distortion to the dataset. We also provide a live evaluation server for submitting intrinsics predictions for the frames in the test split of InFlux++ Real, as well as a live leaderboard to track methods' performance.

For InFlux++ Real, we split the real-world benchmark videos into separate validation and test sets and release the validation set publicly, which contains approximately 15\% of the total frames. For each video, all of its frames belong to the same split. Following InFlux~\cite{influx}, for validation frames, we provide the RGB image and three forms of annotation: ground truth intrinsics without extrapolation, extrapolated intrinsics, and raw lens metadata. For test frames, we only release the RGB image, and users can submit intrinsics predictions for them via our live evaluation server.
\\
\\
For InFlux++ Synth, we release all images and annotations in its entirety. We also release a customizable data loader that allows users to add custom lens distortion and other data augmentations to the images.
\\
\\
Submission instructions are provided through our public GitHub repository and results are reported on a live leaderboard available on our project website. Submissions are evaluated on frames from the test split using only valid ground truth intrinsics, without extrapolation.
\\
\\
The components we release are licensed as follows:
\begin{itemize}
    \item Code: Released under the BSD 3-Clause license, allowing free use, modification, and redistribution.
    \item Dataset and Benchmark Data: Released under Creative Commons Attribution 4.0 (CC BY 4.0). Users may freely use, share, and adapt the dataset, provided that appropriate credit is given and this work is cited in publications.
\end{itemize}
